%% file: archive_vzn.tex
\documentclass{article} %
\usepackage{iclr2027_conference,times}

\input{math_commands.tex}

\usepackage[table]{xcolor}
\usepackage{hyperref}
\usepackage{url}
\usepackage{graphicx}
\usepackage{booktabs}
\usepackage{multirow}
\usepackage{amsmath}
\usepackage{wrapfig}

\graphicspath{{figures/}}

\title{Rethinking Pre-pretraining}

\author{Alexander Stern\textsuperscript{$\dagger$} \\ Cohere \And Tim Rockt\"{a}schel \\ AI Centre, UCL \And Acyr Locatelli\textsuperscript{*} \\ Cohere \And Laura Ruis\textsuperscript{*} \\ MIT}

\iclrfinalcopy
\begin{document}

\maketitle

\begingroup
\renewcommand{\thefootnote}{*}
\footnotetext{Equal contribution.}
\renewcommand{\thefootnote}{$\dagger$}
\footnotetext{Work done partly during a master's at the UCL AI Centre and partly during an internship at Cohere.}
\endgroup

\lhead{Preprint}

\begin{abstract}

We report that allocating compute to synthetic pre-pretraining (PPT) never outperforms pure natural-language training in our compute-matched experiments. This holds across four PPT curricula ($k$-Dyck, neural cellular automata (NCA), declarative reference corpora, and code) and three training regimes: data-constrained, data-unconstrained, and a high token-to-parameter-ratio regime (881M parameters trained on 50B tokens). When natural language is scarce, repeated epochs over a 100M-token corpus beat every synthetic curriculum on natural-language validation loss, with no consistent downstream accuracy advantage from PPT. We find that PPT gains reported in prior work are sensitive to the weight initialization scheme. In an NCA reproduction, we recover a 5.4\% perplexity improvement from PPT under the original initialization scheme. An alternative scheme improves both methods, but the no-PPT baseline more, closing the gap between them. However, when PPT receives \emph{additional compute} rather than a share of the natural-language budget, gains persist at small natural-language budgets even with improved initialization, but disappear by 1B tokens. Finally, models forget what they learn during PPT in subsequent natural-language training; interleaving the same data throughout training prevents forgetting, but does not significantly improve natural-language performance. Our results suggest limited benefits of synthetic pre-pretraining as a distinct training stage and motivate rethinking how such data are used in language model training.

\end{abstract}

\section{Introduction}
\label{sec:intro}

Rich computational structure can be learned from naturally occurring text. The benefits of scaling training on such tokens have yet to hit a ceiling, though increasingly large amounts of data are required for further improvements in performance \citep{kaplan2020scalinglaws}. By contrast, humans acquire language from far less linguistic experience \citep{warstadt-etal-2023-findings}, and their reasoning abilities are largely dissociable from linguistic proficiency \citep{fedorenkoetal2024langnetwork}. This contrast has motivated recent work to ask whether useful linguistic and computational priors can instead be acquired from synthetic data. Beyond making language learning more sample-efficient, synthetic distributions supply effectively unlimited training data once high-quality natural language becomes scarce.

This idea is operationalized in prior work through pre-pretraining (PPT): a training phase on synthetic data that precedes conventional pretraining (PT) and is intended to produce an initialization with useful priors for learning language. Different synthetic distributions target different computational priors: $k$-Dyck encourages hierarchical dependency tracking \citep{hu2025between}, neural cellular automata (NCA) expose models to latent transition rules and long-range dependencies \citep{han2026training}, and procedural tasks train operations such as copying, sorting, and retrieval \citep{jiang2026procedural}. These approaches have reported gains in language modeling and context recall, including up to 8.6\% lower perplexity for NCA, both when PPT is given additional compute on top of a fixed natural-language budget and when PPT and natural-language training share a fixed total compute budget.

\begin{figure}[t]
\centering
\includegraphics[width=\linewidth]{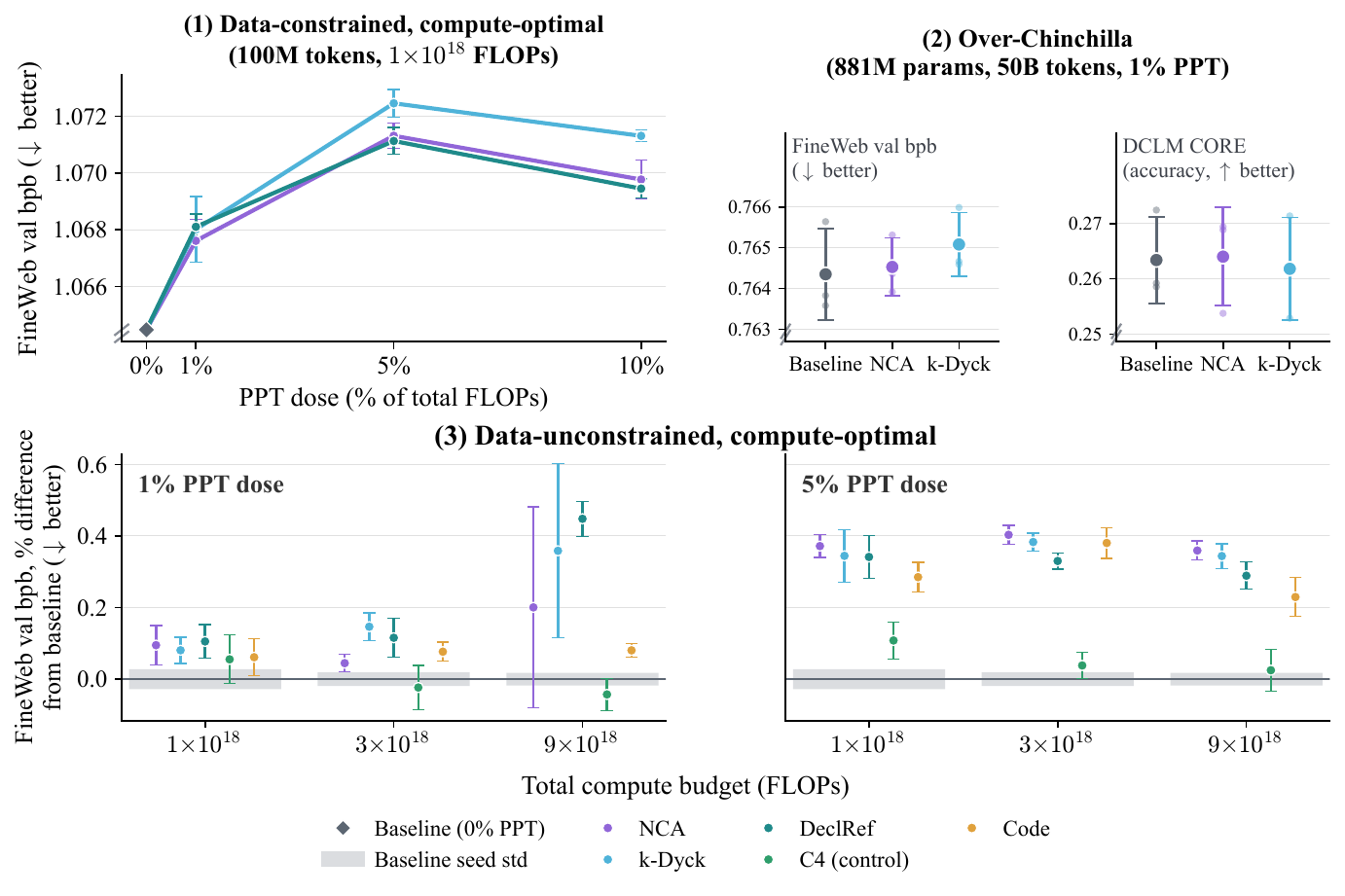}
\caption{\textbf{Across all regimes, allocating compute to PPT never beats
spending it on natural language.}
(1)~Data-constrained: FineWeb validation bits per byte (BPB) against the fraction of a fixed
$1{\times}10^{18}$-FLOP budget spent on PPT, with repetition over 100M natural-language
tokens. Optimal model sizes range from 561M to 708M parameters.
(2)~Over-Chinchilla: an 881M-parameter model trained on 50B tokens,
${\sim}12\times$ the compute-optimal token count, scored on FineWeb BPB and
DCLM CORE. The PPT methods spend 0.5B tokens on PPT and 49.5B on FineWeb.
(3)~Data-unconstrained: the optimal model size found at each budget varies with
the PPT method, from 179M to 186M parameters at $1\times10^{18}$ FLOPs,
252M to 336M at $3\times10^{18}$, and 438M to 561M at $9\times10^{18}$.}
\label{fig:combined_dose_scaling}
\end{figure}

Although prior work has demonstrated gains from synthetic PPT, how robust these gains are across other training configurations and regimes remains unclear. We therefore revisit synthetic PPT across a range of training regimes. We study three synthetic distributions ($k$-Dyck, NCA, and declarative reference corpora (DeclRef)), alongside code, which prior work has found can improve downstream reasoning beyond coding tasks \citep{aryabumi2025to}. 

We begin by asking whether synthetic PPT can provide useful inductive biases when it is not competing with natural-language training for compute. In this additive-compute setting, where PPT is added on top of a fixed natural-language budget, we first reproduce a prior NCA result as a positive control before testing the setting more broadly. We then ask whether PPT is worth its compute when it competes directly with natural-language training for a fixed FLOP budget. We evaluate this compute-matched condition across three regimes: a data-constrained setting, where the alternative is repeated training on the same 100M natural-language tokens; a data-unconstrained setting, which tests whether any PPT recipe remains competitive as training compute scales; and a high token-to-parameter ratio setting, which probes whether PPT becomes useful far beyond the Chinchilla-optimal frontier. Our findings are as follows:

\begin{enumerate}

\item \textbf{Gains from PPT are sensitive to the baseline initialization scheme.} We reproduce a previously reported NCA result under its original setup, recovering a 5.4\% perplexity improvement from PPT. However, changing the token embedding initialization from $\mathcal{N}(0,1)$ to $\mathcal{N}(0,0.02)$ closes the gap between NCA and the no-PPT baseline.

\item \textbf{Additive compute: PPT can help at small natural-language budgets.} When each method is tuned over initialization schemes, $k$-Dyck and NCA improve over the best no-PPT baseline at 50M and 100M natural-language tokens (budgets reported to parallel the amount of language humans may learn from, \citealp{warstadt-etal-2023-findings}). However, by 1B tokens the advantage disappears. This suggests synthetic PPT can instill useful inductive biases, but its benefits diminish at larger natural-language budgets.

\item \textbf{Compute-matched: natural-language training performs as well as or better than PPT across all compute-matched regimes.} This includes the strongest scarcity case, where multiple epochs on 100M unique natural-language tokens beat every synthetic curriculum. The same holds on most of our downstream evaluations, both DCLM CORE \citep{li2025datacomplmsearchgenerationtraining} and BLiMP \citep{warstadt2023blimpbenchmarklinguisticminimal}, with exceptions confined to our smallest compute budget (Table~\ref{tab:aggregate_wins}).

\item \textbf{Mechanism: learned PPT priors may not survive the phase transition to natural language.} We find that gains acquired during PPT on the PPT distribution disappear over the PPT-to-PT phase boundary. Uniformly interleaving the same data throughout training instead preserves performance on those distributions, but we do not find significant improvement in natural-language validation loss.
\end{enumerate}

These findings highlight the surprising effectiveness of naturally occurring data as a training signal: even when unique natural-language tokens are scarce, repeated training on them remains more useful than training on the synthetic corpora we test. Importantly, we do not take this to imply that synthetic data cannot help. Instead, our results point to front-loading synthetic data as a poor way to use it, since what is learned during PPT is progressively lost once natural-language training begins. It may instead be better integrated gradually, through interleaving that avoids the abrupt distribution shift between PPT and PT. Our results therefore motivate rethinking the PPT phase, rather than ruling out synthetic data as a training corpus.

\section{Background \& Related Work}
\label{sec:related}

\subsection{Pre-pretraining}

All prior work in synthetic pre-pretraining shares a common hypothesis: training on certain synthetic distributions can instill language-modeling priors that transfer to natural language. Designed to encourage in-context learning, these distributions can be generated procedurally in effectively unlimited quantities.
PPT corpora in prior work can be divided broadly into three categories: formal languages \citep{hu2025between, mita-etal-2026-language}, algorithmic tasks \citep{wu2022insightspretrainingsimplersynthetic, jiang2026procedural}, and data sampled from randomly initialized neural networks \citep{han2026training, guo2026syntheticprepretrainingimproveslanguage}. Of these, we focus on $k$-Dyck and NCA, for which prior work reports the largest improvements in natural-language perplexity.

\citet{hu2025between} introduce PPT on $k$-Dyck languages, formal languages over a vocabulary of $k$ bracket types whose grammar requires that every opening bracket is eventually closed. For example, at $k=2$, with vocabulary \{\texttt{(}, \texttt{)}, \texttt{[}, \texttt{]}\}, the string \texttt{([()][])} is valid. They find the $k$-Shuffle Dyck variant performs best at $k=128$, and it is this variant we refer to as $k$-Dyck throughout the paper. \citet{han2026training} instead use NCA for PPT. A cellular automaton is a grid of cells, each holding a value from a finite vocabulary---for example, one of ten symbols. The grid evolves over discrete time steps: at each step, every cell takes a new value as a function of the cells in its immediate neighborhood, using a single update rule applied identically to every cell. This can be defined as follows: let $x_t$ denote the state of the whole grid at time step $t$, then an initial state $x_0$ and an update rule $f$ give $x_t = f(x_{t-1})$. An NCA replaces $f$ with a randomly initialized convolutional neural network. 

To construct an NCA PPT corpus, \citet{han2026training} sample a random initial grid, with each cell assigned a value drawn from the fixed vocabulary, together with a randomly initialized convolutional neural network as the update rule. They then run the simulation forward for a fixed number of time steps. At this point, the resulting rollouts range from low-entropy static grids to more complex, visually interesting ones. Hence, each rollout is further filtered by a proxy-complexity metric, the gzip compression ratio, and only those falling within the highest compression band are kept. Each grid in a retained rollout is then flattened exactly as a 2D image tensor would be, and the flattened grids are concatenated in time step order into a one-dimensional token sequence for training.

The experimental protocol is largely shared across the PPT methods in the literature. The natural-language token budget and training recipe are fixed, and a synthetic data phase is prepended, with the number of PPT tokens varied to optimize the downstream natural-language loss after PT. The usual control replaces the synthetic tokens with an equal number of natural-language tokens, and a gain is reported if PPT reaches a lower downstream loss than this control. Methods are therefore compared at equal total tokens and equal model size.

In their largest-scale experiment, training a 1.6B-parameter model, \citet{han2026training} report that PPT on 164M NCA tokens improves perplexity by 5.7\% compared to the no-PPT baseline.
In \citet{hu2025between}'s largest experiment, pre-pretraining a 1B-parameter model on $k$-Dyck before 1.6B natural-language tokens reduces final cross-entropy loss by around 0.6\% (${\sim}2\%$ in perplexity). Both works additionally report gains on public downstream benchmarks: reasoning benchmarks for NCA like BIG-bench Lite \citep{han2026training}, and BLiMP for $k$-Dyck \citep{hu2025between}. We evaluate both fronts in this work, scoring every method on BLiMP \citep{warstadt2023blimpbenchmarklinguisticminimal}, and on the DCLM CORE suite \citep{li2025datacomplmsearchgenerationtraining}, which contains BIG-bench tasks, among them a subset of BIG-bench Lite. A recent study reproduced the $k$-Dyck gain \citep{hu2025between} under the original conditions. However, it found that the gain did not consistently persist across all model sizes, tokenizers, and languages tested \citep{riazhskykh2026instabilityllmprepretrainingdoesnt}. These mixed results motivate our study, and since $k$-Dyck has already been reproduced, our own reproduction (Section~\ref{sec:fixed_nat_lang_add_comp}) focuses on NCA. Concurrently with our work, \citet{yamaguchi2026synthetic} study PPT on $k$-Shuffle Dyck and NCA at larger scales, where they report downstream benchmark gains for PPT methods but do not claim improvements in validation loss. However, like prior work, they also keep the weight-initialization scheme fixed. We find that ablating it significantly improves the natural-language baseline, enough to close its gap with PPT.

\subsection{Training regimes}

How a compute budget should be spent depends on which resource is scarce. When natural language is the constraint, additional compute buys only repeated passes over the same tokens. \citet{muennighoff2023scaling} characterize this setting and find that, for their models and data, up to four epochs of repetition are close to lossless, after which the gains from additional compute rapidly diminish and new data is more useful. The slowrun benchmark \citep{qlabs2026slowrun} isolates the regime in its extreme form, fixing a 100M-token sample of FineWeb and leaving compute essentially unconstrained. Work in this setting has shown that heavy regularization and ensembling continue to buy substantial data efficiency well past the point at which ordinary training overfits \citep{kim2025infinitecompute}. PPT is motivated by the same scarcity, so these are competing answers to one question: what to do with compute once the natural-language data has run out. When data is instead plentiful, compute becomes the constraint, which is the setting where \citet{hoffmann2022training} show that model size and token count should be scaled together. Deployed models, however, are rarely trained at that ratio, because training compute is not the only cost; once inference is accounted for, this moves the optimum toward smaller models trained on more tokens \citep{sardana2024beyond, touvron2023llama}. These are the three compute-matched regimes we study: data-constrained, data-unconstrained, and over-Chinchilla, a comparison that prior PPT work has not explicitly made. Because the compute-optimal model size may itself depend on data quality and curriculum \citep{bi2024deepseek}, in the data-constrained and data-unconstrained regimes we search model size separately for every method and budget.

\section{Experimental Setup}
\label{sec:setup}

The research question we explore is: \textbf{in additive and compute-matched settings, does PPT yield a measurable improvement over a natural-language baseline?} We test this question in four settings, one in the additive setting, followed by three compute-matched configurations. In this section we describe the architecture, datasets, hyperparameter optimization, and evaluation protocol shared across all four settings.

\paragraph{Additive compute.}                                                      
We first evaluate the additive setting, where PPT is prepended to a natural-language      
run rather than substituting any part of its training budget \citep{han2026training, hu2025between, jiang2026procedural}. We fix model size at 336M parameters, and set the natural-language        
PT budgets to  $\{50\mathrm{M},      
100\mathrm{M}, 1\mathrm{B}\}$. Then, for each PT budget, we vary the number of                
PPT tokens over $\{5\mathrm{M}, 20\mathrm{M}, 50\mathrm{M},      
100\mathrm{M}, 200\mathrm{M}, 500\mathrm{M}, 1\mathrm{B}\}$, asking which quantity of PPT minimizes
final natural-language validation loss. We ablate over three weight-initialization schemes: 
the scheme used by \citet{hu2025between}, which initializes every parameter from a Normal distribution $\mathcal{N}(0, 0.02)$; the scheme used by \citet{han2026training}, which            
initializes token embeddings from the standard Normal distribution $\mathcal{N}(0, 1)$ and all remaining parameters   
from $\mathcal{N}(0, 0.02)$; and the spectral-based initialization of NanoChat used throughout the     
rest of this paper \citep{yang2024spectralconditionfeaturelearning, nanochat}, which draws linear-layer weights from $\mathcal{N}(0,\sigma)$ with $\sigma = \min(1,\sqrt{d_{\text{out}}/d_{\text{in}}})/\sqrt{d_{\text{in}}}$, token embeddings from $\mathcal{N}(0,1)$, and zero-initializes the language-modeling head and all residual projection weights. This ablation is additional to prior work, and turns out to be crucial for answering our research question. 

\paragraph{Compute-matched.}
For the rest of the experiments, we instead fix the total compute budget and ask whether PPT can justify a share of it. Each experiment begins from a total training budget $C$, measured in FLOPs, and $\alpha$ giving the fraction of that budget spent on PPT. A run with $\alpha$ receives $\alpha C$ FLOPs of
PPT followed by $(1-\alpha)C$ FLOPs of natural-language pretraining. Hence, a run with $\alpha = 0$ spends the entire budget on natural language. We use
$\alpha \in \{0, 0.01, 0.05, 0.10\}$, with the largest percentage used only in the
data-constrained regime: early experiments at the larger budgets showed no benefit
that justified the extra runs, so $\alpha = 0.10$ was not run there.
Note that $\alpha$ is a fraction of compute, not of tokens: the number of tokens
each phase consumes follows from the allocation rather than being set directly. Our FLOP measurement method is covered in Appendix~\ref{app:flop_measure}.

The compute-matched experiments span three configurations. The                     
\emph{data-constrained} configuration fixes natural language at 100M tokens and the  
total budget at $1\times10^{18}$ FLOPs. The \emph{data-unconstrained} configuration leaves natural language abundant and sweeps over $1\times10^{18}$, $3\times10^{18}$ and $9\times10^{18}$ FLOP budgets. At each budget, we search over model size to minimize validation loss so models are trained at compute-optimality. The \emph{over-Chinchilla} configuration trains a model of 881M parameters on 50B tokens, ${\sim}57$ tokens per parameter, evaluating NCA and $k$-Dyck, the two curricula from prior work we revisit, each at 1\% (500M tokens). In all three compute-matched settings we use only the spectral-based default initialization of the NanoChat version we build on \citep{nanochat}, as it has been well tuned by the NanoChat community. Using this initialization is further supported by our additive compute experiments, where at the largest natural-language budget the spectral-based initialization performed best for both the baseline and the PPT methods (Figure~\ref{fig:token_dose}).

\begin{figure}[t]
\centering
\includegraphics[width=\linewidth]{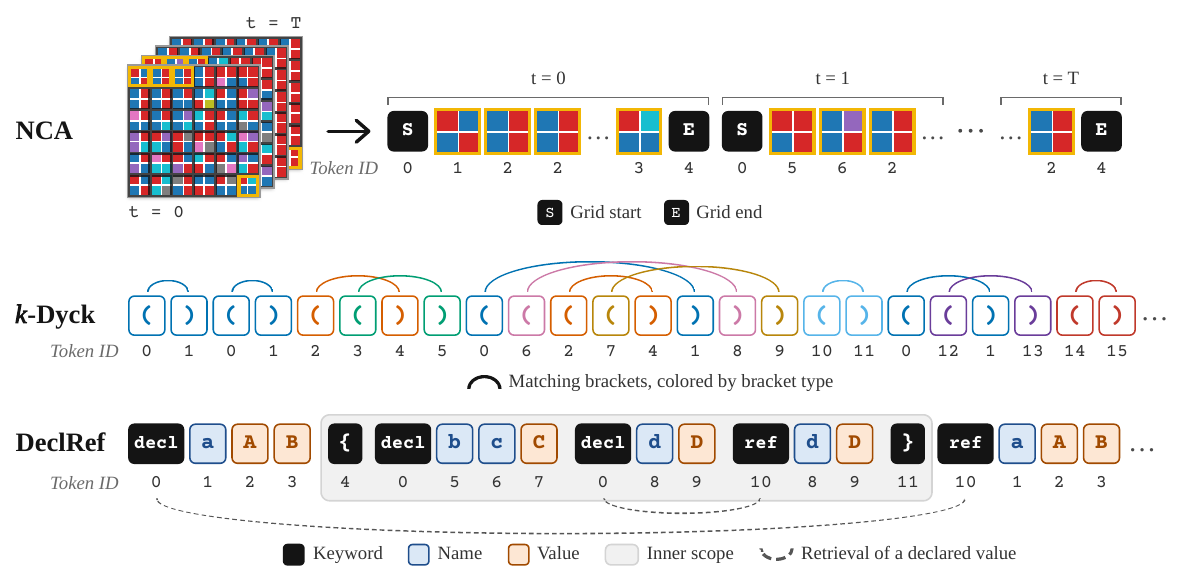}
\caption{\textbf{Tokenizing synthetic pre-pretraining data.} For NCA, each grid of a rollout is split into $2\times2$ patches that each become one token, and grids are concatenated in time order with start and end tokens. For $k$-Dyck, the opening and closing bracket types are each their own token. For DeclRef, keywords, scope brackets, names and values are unique tokens. In the example shown, a reference inside the opened scope retrieves the value of a name declared in that scope, and after the scope closes a reference retrieves the value of a name declared before the scope. Generation and tokenization details are given in Section~\ref{sec:related} and Appendix~\ref{app:data}.}
\label{fig:ppt_data}
\end{figure}

\paragraph{Architecture and data.}
All our models are trained using a fork of NanoChat, with the Muon optimizer \citep{jordan2024muon, liu2025muonscalable, nanochat}, which updates the transformer blocks' weight matrices, while the embedding and unembedding tables are trained under AdamW.
However, we find that in our smaller-scale experiments the PPT phase often does better under AdamW, while at larger scales Muon performs on par, and so we ablate AdamW, applied to all parameters, only for the additive-compute experiments and the $1{\times}10^{18}$ data-constrained experiment. NanoChat sets the hidden size to $64d$ for a model of depth $d$, so the parameter count follows from the depth alone. Depth is therefore the only capacity-related hyperparameter we search.

For our main natural-language corpus we follow NanoChat and use FineWeb \citep{penedo2024fineweb}. For $k$-Dyck and NCA, we generate the data using the same code released by prior work (Appendices~\ref{app:kdyck} and~\ref{app:nca_corpus}). We additionally introduce DeclRef, a synthetic corpus that, like the procedural tasks of \citet{jiang2026procedural}, targets retrieval from context (Appendix~\ref{app:declref}). Examples of these PPT distributions can be seen in Figure~\ref{fig:ppt_data}. Alongside the synthetic curricula, we pre-pretrain on two real-text corpora, C4 \citep{raffel2023exploringlimitstransferlearning} as a natural-language control and the GitHub Code dataset \citep{codeparrot2022githubcode}. The additive setting is aimed at revisiting the additive-compute claims made for the synthetic distributions of prior work, NCA and $k$-Dyck, so for this setting we do not test DeclRef. Our data-constrained experiments add DeclRef but leave out code and C4, since these supply additional real text, which would break the 100M-token constraint on natural language. The data-unconstrained experiments do include these other corpora, while the over-Chinchilla experiments return to NCA and $k$-Dyck alone. Following prior work \citep{han2026training}, the synthetic languages are encoded with their own, far smaller vocabularies. At the phase boundary we therefore expand the vocabulary to its natural-language size and reinitialize both the token embedding table and the language-modeling head, carrying only the transformer body across; each phase has its own learning-rate schedule and the optimizer state is reset at the boundary. Code and C4 share the natural-language tokenizer, so their embedding table and head carry over unchanged. The C4 control instead runs both its PPT and PT phases under a single learning-rate schedule with no optimizer state reset, since in our experiments resetting the optimizer state at the boundary hurt its final loss (Appendix~\ref{app:search_config}). This deviates from prior implementations, which run the natural-language control as a separate phase with its own learning-rate schedule and an optimizer state reset \citep{hu2025between, han2026training, yamaguchi2026synthetic}. We recommend this setup for natural-language controls in future PPT work.

\paragraph{Hyperparameter optimization.}                                              
A subset of our hyperparameters are optimized with \texttt{LLMTuna}, a language-model-based       
hyperparameter optimizer we release alongside this paper. \texttt{LLMTuna} optimizes held-out validation bits per byte (BPB) by proposing each    
successive hyperparameter configuration from a rolling context of prior trials and their          
results, together with any files in the codebase one may want to include as context; this is in line with recent work     
using language models as hyperparameter optimizers \citep{yang2023large, jiang2026autoresearch, karpathy2026autoresearch}.  
We use it because we found it more sample-efficient than Bayesian hyperparameter optimization using Optuna \citep{akiba2019optuna} (see Appendix~\ref{app:hpo_compare}). The hyperparameters searched and held fixed in each setting are detailed in Appendix~\ref{app:hparams}.

\paragraph{Evaluation.}
Our primary metric is BPB on a held-out split of FineWeb ($\sim$10M tokens); hyperparameter tuning uses a separate validation set. We additionally evaluate on held-out splits of C4 and GitHub code (Appendix~\ref{app:extra}), on the DCLM CORE suite \citep{li2025datacomplmsearchgenerationtraining}, and on BLiMP \citep{warstadt2023blimpbenchmarklinguisticminimal}, the grammaticality benchmark used by \citet{hu2025between}. DCLM CORE aggregates 22 benchmarks covering knowledge, commonsense, reading comprehension, and symbolic reasoning (Appendix~\ref{app:benchmarks}). Every metric is reported as the mean over three seeds, with $\pm$ values and error bars giving one sample standard deviation.

\begin{figure}[t]
\centering
\begin{minipage}[c]{0.48\linewidth}
\centering
\includegraphics[width=\linewidth]{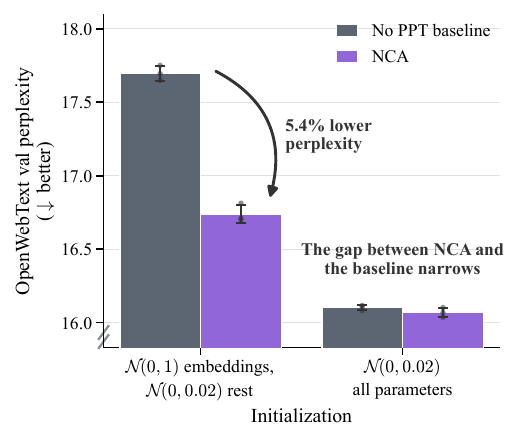}
\end{minipage}\hfill
\begin{minipage}[c]{0.48\linewidth}
\caption{\textbf{Gains from NCA PPT are sensitive to baseline
initialization.}
A 400M-parameter Llama model trained on ${\sim}9$B tokens of OpenWebText, reproducing an NCA PPT experiment of \citet{han2026training}; we plot perplexity to match how prior work reports it. The groups differ only in weight initialization. With $\mathcal{N}(0,1)$ embeddings and $\mathcal{N}(0,0.02)$ elsewhere, NCA PPT improves perplexity over the no-PPT baseline by 5.4\%. Initializing the embeddings from $\mathcal{N}(0,0.02)$ instead improves the baseline itself by 9.0\%, and the two methods become inseparable despite the additional compute the NCA phase receives.}
\label{fig:nca_init}
\end{minipage}
\end{figure}

\section{Additive Compute}
\label{sec:fixed_nat_lang_add_comp}

\textbf{The reproduced NCA advantage vanishes under an alternative initialization.} We begin by reproducing one of the experimental setups of \citet{han2026training}, from their \emph{Section~5.1}. The exact setup is not described in Section~\ref{sec:setup}, but fits within the additive-compute regime. We train a 400M-parameter Llama model on ${\sim}9$B tokens of
OpenWebText, and PPT on 164M NCA tokens. Doing so delivers a 5.4\% perplexity improvement over the no-PPT
baseline. However, changing only the scale at which the embeddings are initialized \citep{hf2024llamaconfig} improves the baseline by 9.0\%---more than the reported gain itself---and applying the same initialization to the NCA method leaves the two indistinguishable (Figure~\ref{fig:nca_init}). For more details on the reproduction see Appendix~\ref{app:nca_repro}. Note that this experiment isolates sensitivity to initialization, but does not identify the best NCA configuration under the alternative initialization, since we do not retune the number of PPT tokens in that setting. However, this experiment is meant to show how much performance gain is available on the baseline before any PPT data is introduced, rather than to establish the best achievable NCA configuration under the original training scheme. The latter is studied in the additive-compute and compute-matched experiments that follow.

\begin{figure}[t]
\centering
\includegraphics[width=\linewidth]{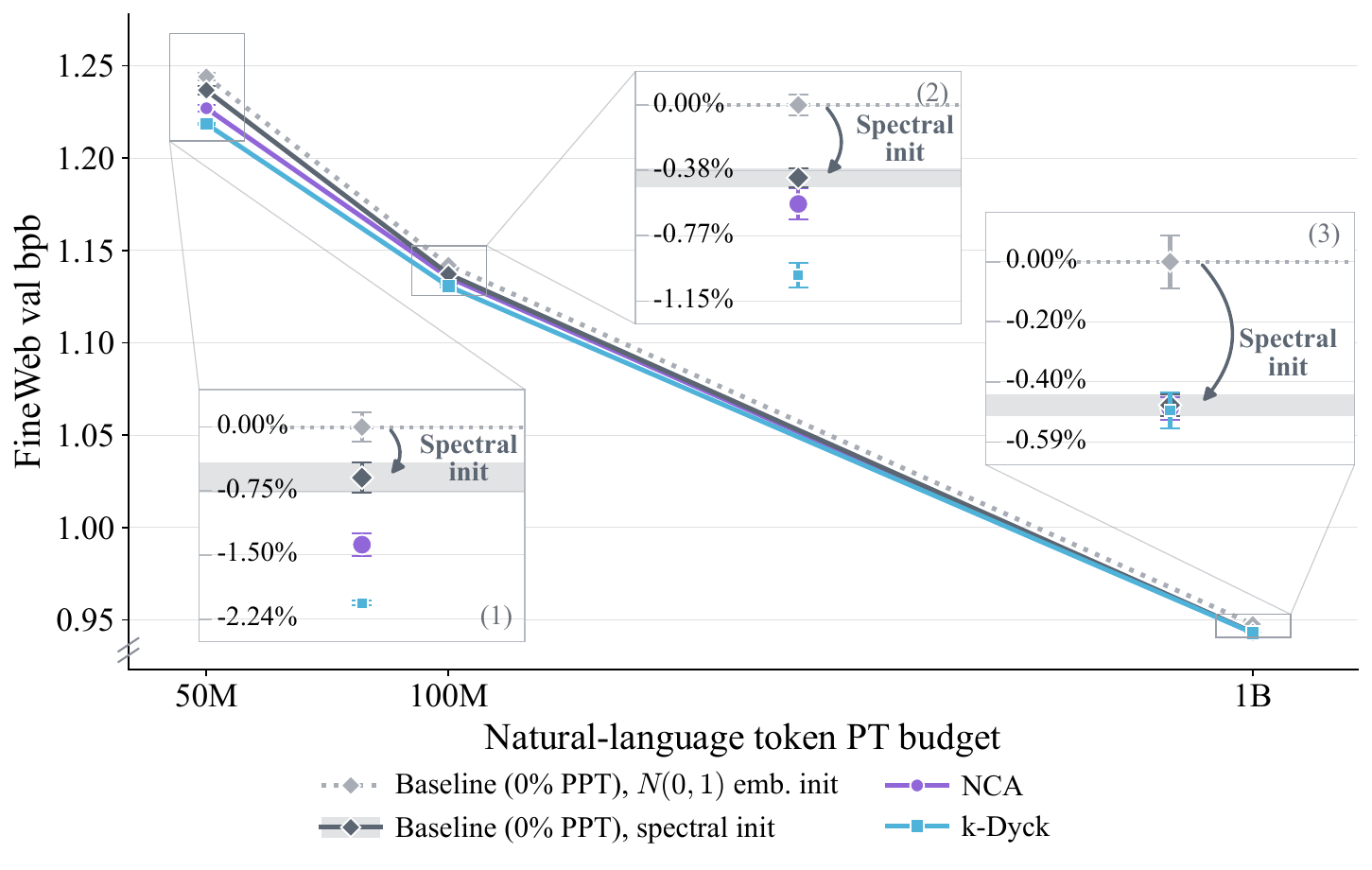}
\caption{\textbf{PPT is useful at 50M--100M natural-language tokens, but its gain at 1B disappears under spectral-based initialization.} In the additive-compute regime with a 336M-parameter model, NCA and $k$-Dyck, each at its lowest-loss configuration, improve on the optimal no-PPT baseline at 50M and 100M natural-language tokens. Under spectral-based initialization this gain disappears by 1B tokens, but under the other initializations it persists at every budget (Appendix~\ref{app:additive_full} shows the full breakdown). Because spectral-based initialization yields the lowest loss for every method at this largest budget, we use it in all compute-matched experiments, which run at comparable FLOP budgets.}
\label{fig:token_dose}
\end{figure}

\textbf{PPT helps at small natural-language budgets, but not at larger ones.} We next ask whether PPT can help in the additive setting, where the natural-language budget is tightly constrained and the synthetic phase does not compete for it. Figure~\ref{fig:token_dose} shows, for each PPT method at each natural-language token budget, the best PPT token count and weight initialization. At 50M and 100M natural-language tokens both PPT methods improve over the best no-PPT baseline, whose winning configuration uses spectral-based initialization. This means that at these low budgets the synthetic languages do instill useful inductive biases. This is also seen in Table~\ref{tab:aggregate_wins}, where $k$-Dyck PPT at the 50M natural-language token budget improves DCLM CORE accuracy over the no-PPT baseline. However, at 1B tokens, under the spectral-based initialization, all PPT methods fall within the baseline’s seed spread. Yet under the alternative baseline initialization, both PPT curricula still appear to improve over the baseline at the 1B token scale. The additive setting therefore shows two overlaid effects: at small natural-language budgets, PPT helps regardless of initialization, whereas at 1B it only helps against baselines without spectral-based initialization. For further details on these additive-compute experiments, see Appendix~\ref{app:additive_full}, where Figure~\ref{fig:token_dose_curves} breaks the results down by initialization and PPT optimizer.

\section{Compute-Matched}
\label{sec:constrained}

Although the previous section shows that PPT can improve performance when compute is added on top of a fixed natural-language budget, this does not establish that synthetic data is the best use of those resources. In practice, the same compute could instead be spent on further natural-language training, including repeated epochs when unique language is scarce. We therefore turn to compute-matched experiments, asking whether PPT can justify diverting part of the training budget away from natural language.

\begin{table}[h]
\centering
\small
\setlength{\tabcolsep}{4pt}
\input{tables/aggregate_wins}
\caption{\textbf{Downstream benchmark improvement.} Across the models reported in Figures~\ref{fig:combined_dose_scaling} and~\ref{fig:token_dose}, these are the cases whose aggregate BLiMP or DCLM CORE score sits above their baseline. However, C4, the natural-language control, reaches a DCLM CORE score in the same range as the PPT methods. In this table, NL denotes natural-language tokens. }
\label{tab:aggregate_wins}
\end{table}

\textbf{Across the tested regimes, allocating compute to synthetic PPT never beats spending it on natural language.} Figure~\ref{fig:combined_dose_scaling} shows results across the three compute-matched configurations. In the data-constrained regime, diverting any fraction of the budget to synthetic PPT increases validation BPB, with the effect leveling off beyond 5\%.  The compute is better spent on additional epochs over the same natural-language data. In the over-Chinchilla regime, neither NCA nor $k$-Dyck improves over the baseline on validation loss or DCLM CORE accuracy. The data-unconstrained scaling confirms this pattern holds across budgets: no PPT method significantly outperforms the baseline in terms of natural-language validation BPB, with 1\% closer to the baseline than 5\% at the two smaller budgets, and the natural-language C4 control matching the baseline. To check this trend, we fit a power law to each method's results across these budgets and find that no PPT method has either a lower multiplicative constant or a significantly larger scaling exponent than the baseline (Appendix~\ref{app:scaling_unconstrained}, Figure~\ref{fig:scaling_fits}). We therefore find no evidence that these recipes scale better than the baseline. The BPB results are broadly reflected in the downstream DCLM CORE and BLiMP benchmarks, with the exceptions shown in Table~\ref{tab:aggregate_wins}. At the lowest compute budget we study, DeclRef and code show some improvement over the baseline. However, how significant these gains are remains unclear. PPT on C4, the natural-language control, scores similarly to these synthetic PPT methods on DCLM CORE. In addition, these are the only experiments in which we observe an aggregate significant gain; every other method performs at or below the baseline (see Appendix~\ref{app:extra}). However, when breaking the two benchmarks into their individual tasks, we find that some PPT methods do improve on specific sub-tasks, but often at the cost of others, which is why most perform on par on aggregate (Appendix~\ref{app:benchmarks}, Figure~\ref{fig:subscore_direction}). This is especially the case for BLiMP, where some phenomena improve under PPT, in line with the BLiMP gains \citet{hu2025between} report for $k$-Dyck.

We also observe no consistent ranking among PPT methods across configurations or scales. For example, DeclRef is marginally better than NCA and $k$-Dyck in the data-constrained regime, but in the data-unconstrained regime the ordering varies with both the PPT percentage and the total compute budget. This is accentuated in the over-Chinchilla regime: with large token budgets and parameter counts, final validation loss between the baseline and PPT methods becomes very similar.

\section{Retention Across the PPT-to-PT Transition}
\label{sec:forgetting}

\begin{figure}[h]
\centering
\includegraphics[width=\linewidth]{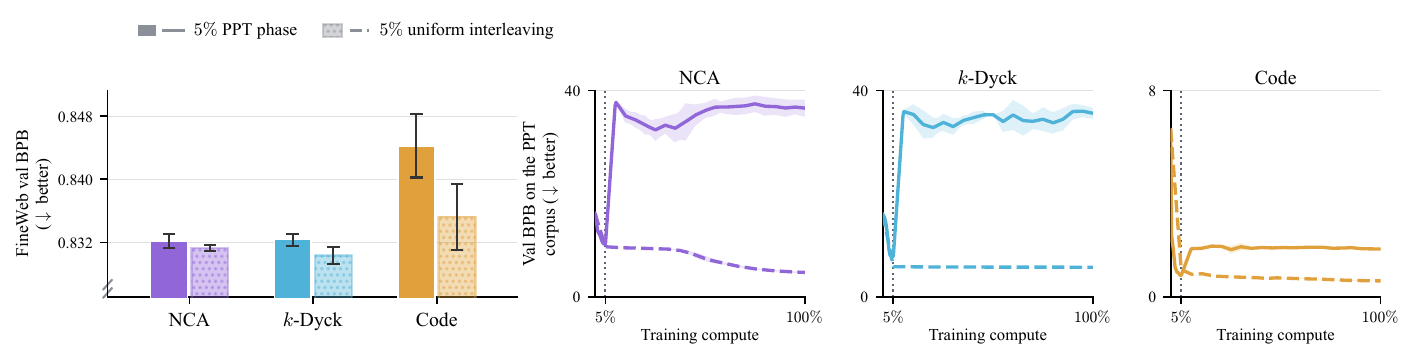}
\caption{\textbf{Uniformly distributing the PPT data across training, rather than front-loading
it, prevents forgetting on the PPT distribution.}
In the interleaved runs, $5\%$ of every batch consists of NCA, $k$-Dyck, or
code throughout training. In the PPT runs, the same $5\%$ of compute is instead
spent up front, at $2.7\times10^{19}$ FLOPs. Left: FineWeb validation BPB at the
end of training. Right: validation BPB on the PPT corpus itself across the whole run, one
panel per corpus on its own scale, with the gray dotted line marking $5\%$ of total compute.
Under the PPT schedule the models lose this ability at the PPT-to-PT boundary; under
interleaving they keep it and continue to improve on it.
Elsewhere in this work the vocabulary is replaced at the phase boundary for the
synthetic PPT distributions, which means the PPT corpus cannot be scored after the
PPT phase. Every run here instead shares a single vocabulary, extended to hold
the NCA and $k$-Dyck tokens, so the corpus can be tracked throughout training.
These runs are not hyperparameter-tuned (configurations in Appendix~\ref{app:hparams_found}, Table~\ref{tab:hparams_interleave}), so the FineWeb panel is included only to show that the interleaved runs sit in a similar range to their PPT counterparts.}
\label{fig:interleave}
\end{figure}

One reason PPT may fail to improve over the baseline is that what it learns does not carry over into PT. The shift from PPT to PT is abrupt: the PPT data differs substantially from natural language, and is seen only at the start of training and never again. \citet{ibrahim2024simple} show that even a shift between two natural languages causes forgetting. Their work shows that continued pretraining on German, after 300B tokens of English, causes forgetting on the original English distribution. The shift from a synthetic distribution to natural language is even larger.

We test this hypothesis by tracking the validation BPB of the NCA, $k$-Dyck and code PPT distributions across the phase boundary, at $2.7\times10^{19}$ FLOPs, a larger budget than the ones run in previous sections. This gives us a longer horizon over which to observe performance on the PPT distribution. The three right-hand panels in Figure~\ref{fig:interleave} show the result: during the PPT phase all three methods learn their respective distributions, but at the PPT-to-PT boundary the BPB immediately starts increasing, signaling that the models forget their ability on the PPT distribution. The same figure suggests that this stems from front-loading the data in the PPT phase. Holding 5\% of every batch as PPT data throughout training instead improves performance on the PPT distribution over the entire run. This is despite both schedules spending the same total compute on the PPT distribution. Forgetting at the boundary is therefore preventable by simply mixing in the synthetic distribution throughout PT. Interleaving improves FineWeb BPB for the PPT methods, though not significantly for all, and as these runs are not hyperparameter-tuned at this scale, how large the gain could be remains open.

\section{Limitations}

\paragraph{Hyperparameter search at larger scales.} We could not run a full hyperparameter search above $9\times10^{18}$ FLOPs, where each trial becomes expensive; the over-Chinchilla runs, for instance, sweep only the peak learning rates and use other hyperparameters from the smaller budgets (Appendix~\ref{app:overchinchilla}). However, the baseline and PPT methods receive the same reduced sweep, so the comparison between them remains matched. Similarly, the share of compute given to PPT is itself a hyperparameter, which we fix at 1\% and 5\% in the data-unconstrained setting rather than tune, leaving shares in between, or ones that change with the total budget, untested.

\paragraph{Effect sizes relative to prior work.} The differences we report between PPT and the baseline are small, and in some settings smaller than those previously reported. One reason is that we use NanoChat, which uses a different tokenizer, model architecture, and natural-language dataset than prior work. The tokenizer alone has been shown to change the effect of PPT \citep{riazhskykh2026instabilityllmprepretrainingdoesnt}. However, within our own experiments, the setup is held fixed across the baseline and the PPT methods, so any difference between them is attributable to PPT rather than to these details. Another reason for discrepancies with prior gains is the metric we report. In order to make our results easily comparable for future work, we evaluate using validation BPB. Prior work often reports results in perplexity, which is the exponential of the loss and hence yields larger relative differences than BPB. Hence, our BPB differences are not directly comparable to the perplexity ones shown in prior work.

\section{Discussion}
\label{sec:conclusion_limitstions}

We set out to test whether PPT improves PT efficiency. Across three synthetic corpora and multiple training regimes, allocating compute to PPT never outperforms spending it on natural language. This holds even under a multi-epoch setting over 100M tokens. However, in the additive setting at the smallest natural-language budgets, a benefit persists even against a well-initialized baseline. Prior studies hold the initialization scheme fixed while varying the curriculum, a natural design choice for isolating the curriculum effect. Our results suggest that future robustness checks should additionally vary this axis for both PPT and no-PPT methods.

Previous work argues that the priors learned during PPT persist through to the PT phase \citep{hu2025between, han2026training}. In our experiments, however, models do not retain their ability on the PPT distribution after the transition (Section~\ref{sec:forgetting}). If such priors are indeed kept after PPT, we would expect the model to keep at least part of this ability. One could argue that losing this ability is expected, and that it may not matter how well the model does on the PPT distribution, as long as the priors survive. Once the PPT data is no longer present in training, the model no longer needs to assign probability to the PPT tokens, so it is likely that the performance on the PPT distribution degrades. However, if this were the sole reason for the drop in performance, degradation on the code distribution would not occur. Unlike the synthetic distributions, most tokens seen during the code PPT phase are also seen during the PT phase, since comments and variable names carry natural-language semantics. Hence, the model keeps learning to predict them during PT, albeit outside of code contexts. Yet, even in this case, loss on the code distribution rises after the phase boundary. This suggests that the priors learned on code-specific data are not fully kept, and are instead progressively overwritten once the distribution shifts abruptly to natural language. We therefore recommend that future work on synthetic data schedules also measure the model's ability on the synthetic distribution throughout training, as a proxy for whether the inductive biases learned from it are retained.

A natural question is whether PPT could pay off at scales beyond those tested in this work. Though our largest models stay below one billion parameters, our over-Chinchilla experiments train an 881M-parameter model on 50B tokens, a compute budget comparable to or larger than experiments in prior PPT work \citep{hu2025between, han2026training, yamaguchi2026synthetic}. Even so, we show that neither NCA nor $k$-Dyck improves on the baseline at this scale. We also approach the question from a scaling-law perspective, which motivates our data-unconstrained experiments. Such a perspective has not been explicitly analyzed in prior work. There, we search for each method's compute-optimal recipe at three budgets and fit a power law to each, and find no evidence that any PPT method scales better than the baseline (Section~\ref{sec:constrained}, Appendix~\ref{app:scaling_unconstrained}). Taken together, these results suggest that these PPT recipes would not overtake natural-language training at larger scales.

Next, our main conclusions concern natural-language validation loss. On downstream benchmarks, where prior work highlights gains \citep{hu2025between, han2026training, yamaguchi2026synthetic}, for example on specific BLiMP phenomena \citep{hu2025between} and on reasoning benchmarks after fine-tuning \citep{han2026training}, the picture is less clear. In our experiments, aggregate BLiMP and DCLM CORE scores mostly match the baseline. However, exceptions exist on individual sub-tasks, including BLiMP phenomena, which shift in both directions, though without a clear pattern across methods or settings (Figure~\ref{fig:subscore_direction}). Hence, PPT may still shape narrower capabilities that aggregate benchmarks do not capture, which more targeted evaluations could bring out. Where it does, integrating synthetic data gradually rather than front-loading it could help better retain these capabilities. PPT could similarly prove more useful in other data modalities, which our study, limited to language modeling, does not examine.

Moreover, our findings leave open questions for future work. As natural-language data is increasingly constrained relative to available compute \citep{kim2025infinitecompute}, synthetic data is a compelling prospect. That said, our results motivate rethinking whether such data should be integrated via an explicit PPT phase. For such a phase to be worthwhile, the forgetting across the phase boundary must first be solved. Alternatively, this forgetting can be avoided by interleaving synthetic data throughout training. We therefore recommend that future work investigate better interleaving schedules. However, whether such schedules make synthetic data a better use of compute than natural language remains an open question.

\section*{Acknowledgements}

We thank Dan Lee and Seungwook Han, authors of \citet{han2026training}, and Michael Hu, author of \citet{hu2025between}, for helpful discussions and advice on this work. Specifically, Dan Lee and Seungwook Han suggested comparing AdamW with Muon for the PPT phase in our additive experiments, which improved the PPT methods (Figure~\ref{fig:token_dose_curves}).

\section*{Reproducibility Statement}

Links to our code are given in Appendix~\ref{app:code}. The repositories contain the full training factory, the data-preparation and generation scripts for every corpus, and the evaluation code, together with a manifest holding one record per experiment we report.

\section*{AI Use Statement}

In this work, we used generative AI tools to help write the code in the repositories we release (Appendix~\ref{app:code}), find related work, and polish the writing of this paper, all under close supervision by us, who reviewed and edited all AI-assisted output. In addition, our hyperparameter optimizer, \texttt{LLMTuna}, uses a language model to propose hyperparameter configurations within a purpose-built search loop and not simply a coding agent (Appendix~\ref{app:LLMTuna}). We take responsibility for the final content of this work, including text, claims or artifacts produced with the aid of generative AI.

\bibliography{archive_vzn}
\bibliographystyle{iclr2027_conference}

\appendix

\section{Data}
\label{app:data}

\subsection{$k$-Dyck}
\label{app:kdyck}

We generate $k$-Shuffle Dyck with the code released by \citet{hu2025between}, at
their published settings. At each position the generator opens a new bracket
with probability $0.5$ or closes one of the currently open brackets, chosen
uniformly among the open types, forcing a close once the maximum nesting depth
is reached and an open at depth zero. We use the variant and setting they report
as best, $k=128$ bracket types, which gives a vocabulary of $2k=256$, with a
maximum nesting depth of $16$ and sequences of $2048$ tokens. 

\subsection{Neural cellular automata}
\label{app:nca_corpus}

We generate the NCA corpus with the code released by \citet{han2026training}.
The automaton is a $12\times12$ grid of cells, each holding one of ten colors,
and the update rule is a randomly initialized convolutional network applied
identically to every cell. A rule and an initial grid are sampled for every
sequence, and a rule is kept only when the gzip compression ratio of its rollout
exceeds $50\%$, which selects for dynamics that are more complex rather than
trivially predictable (low entropy).

Grids are not tokenized cell by cell. Each grid is serialized in row-major order
as non-overlapping $2\times2$ patches, so its $144$ cells become $36$ tokens,
and since a patch holds four cells over ten colors the patch vocabulary is
$10^4$ token IDs. A start and an end token delimit each grid, so one grid costs
$38$ tokens, and successive grids are concatenated into a single sequence
tracing the automaton's evolution.

The context length therefore fixes how many grids a sequence holds.
\citet{han2026training} train at $1024$ tokens, which is $27$ grids
($27\times38 = 1026$, with the tail truncated), whereas we train at $2048$,
which is $54$ grids ($54\times38 = 2052$, again with its tail truncated). Relative to a sequence length of 1024, each of our sequences shows twice as
much of the automaton's evolution as a result. This
does not apply to the NCA reproduction of Appendix~\ref{app:nca_repro}. 

\subsection{Declarative-reference data}
  \label{app:declref}

Prior work uses procedural tasks such as copying and retrieval as PPT data
\citep{jiang2026procedural}. This ``declarative-reference'' corpus follows that idea, aiming at a
distribution with little semantics that forces context recall, but at a higher
entropy than a pure copying task.

In this synthetic corpus, each training document is a sequence of statements, built from four keywords:

  \begin{itemize}
  \item \texttt{declare} declares a new \emph{Name} with its \emph{Value}.
  \item \texttt{reference} repeats a \emph{Name} that is already in scope (i.e., that has been declared),
  followed by its \emph{Value}.
    \item \texttt{open} starts a new \emph{Scope}, and \texttt{close} ends it. A Scope is the part of a document between a matching \texttt{open} and \texttt{close}. Scopes can be nested, like balanced brackets in $k$-Dyck. Names declared inside a Scope are available until that Scope is closed; Names declared in enclosing Scopes remain available.

\end{itemize}

  Here are the first sixteen tokens of a training document, with token IDs above
  and a possible decoding below. The choice of letters is arbitrary and only
  makes the example more readable: lowercase letters stand for \emph{Name}
  tokens, uppercase for \emph{Value} tokens.

  \begin{center}
  \resizebox{\linewidth}{!}{%
  \begin{tabular}{l|cccccccccccccccc}
  token IDs & 2 & 163 & 873 & 1027 & 977 & 2 & 305 & 383 & 73 & 747 & 730 & 3 & 163 & 873 & 1027 & 977 \\
  \midrule
  decoded & \texttt{declare} & r & Q & S & T & \texttt{declare} & n & b & m & U & V & \texttt{reference} & r & Q & S & T \\
  \end{tabular}}
  \end{center}

  Reading it as three statements:

\begin{enumerate}
\item \texttt{declare r Q S T}. At the start of a document nothing has been
declared yet, so a \texttt{reference} is impossible and the generator emits a
\texttt{declare}. It then draws a random \emph{Name}, here the single token
$r$, and a random \emph{Value}, here the three tokens $Q\,S\,T$. The length of
each is itself drawn at random, from one to three tokens, so statements vary in
length.
\item \texttt{declare n b m U V}. A \emph{Name} is now in scope, so this time
the generator could have emitted a \texttt{reference}. The keyword is chosen at
random from those available, and here it is a \texttt{declare} again, giving
the \emph{Name} $n\,b\,m$ and the \emph{Value} $U\,V$.
\item \texttt{reference r Q S T}. This time the keyword drawn is a \texttt{reference}. A second random choice picks a \emph{Name} from those in scope, either $r$ or $n\,b\,m$, and here picks $r$. Once both are made, the most likely continuation is the \emph{Value} $Q\,S\,T$ associated with $r$, so a model predicts these three tokens correctly by recalling that pairing from the context.
\end{enumerate}
The \emph{Names} and \emph{Values} occupy disjoint blocks of the vocabulary, IDs $[4, 516)$ and $[516, 1028)$, and the four keywords occupy IDs $0$ to $3$, so the letter case in the decoding stands in for a split that is real in the data. No \texttt{open} or \texttt{close} appears in this example window, although both occur elsewhere in the document. When a Scope is closed, Names declared inside it are no longer available for \texttt{reference}; Names declared in enclosing Scopes remain available.

The generator has the following parameters: 

\begin{itemize}
\item How often a \emph{Scope} opens, and how deeply \emph{Scopes} can nest.
\item How frequently each \emph{Name} and \emph{Value} token is drawn. The
frequencies follow a Zipf distribution drawn newly for every document.
\item How a \texttt{reference} chooses among the \emph{Names} currently in
scope, which favors recently declared ones.
\item How the mix of \texttt{declare} to \texttt{reference} statements shifts
from the start of a document to its end.
\item The fraction of references that pair the \emph{Name} with a fresh
\emph{Value} instead of repeating the original one, so that no position is
perfectly deterministic.
\item How often the \emph{Name} frequencies are reshuffled part-way through a document: a \emph{Name} that occurs every few statements early on may, after the reshuffle, become as rare as any other, so frequency statistics inferred from the opening tokens go stale and must be re-estimated from later context.
\end{itemize}

We choose these parameters with the following goal in mind: that predicting a token
should mainly require using the current document rather than memorizing. \emph{Name} and \emph{Value} tokens are drawn partly from a fixed table of likely successors shared by every document, as in natural language, where some word sequences are common across all text, and partly at random. The shared part can be learned across documents, while the random part cannot. The random part aims to provide the corpus with entropy. A \emph{Value} is therefore only partly predictable when it is declared, so a \texttt{reference} must mainly retrieve it from the current document. The Zipf token frequencies are also redrawn for every document, so they must be inferred from the current one. Which \emph{Names} can be referenced at all depends on the \emph{Scopes} currently open, so both the answer and the space of possible answers are properties of the local context. The parameter values are provided in the generator code released with this paper.

\section{NCA Reproduction Experiment}
\label{app:nca_repro}

We follow one of the configurations described by \citet{han2026training}, a 400M-parameter Llama model
trained on ${\sim}9$B tokens of OpenWebText, for which the original work reports
an 8.6\% perplexity improvement over the no-PPT baseline. The exact Llama
configuration is not specified for the 400M model, so we estimate it and choose
the following: 24 layers, a hidden size of 1024 with 16 attention heads, and a
multilayer perceptron intermediate size of 4096. The OpenWebText training and validation splits
are not specified either, so we choose these ourselves too. Under their
initialization we recover the effect, a 5.4\% gain, and we attribute the
remaining difference from the reported 8.6\% to the choices above. Beyond those
choices we make no changes to the released codebase, and the only thing we vary
is the weight initialization.

\begin{figure}[h]
\centering
\includegraphics[width=0.75\linewidth]{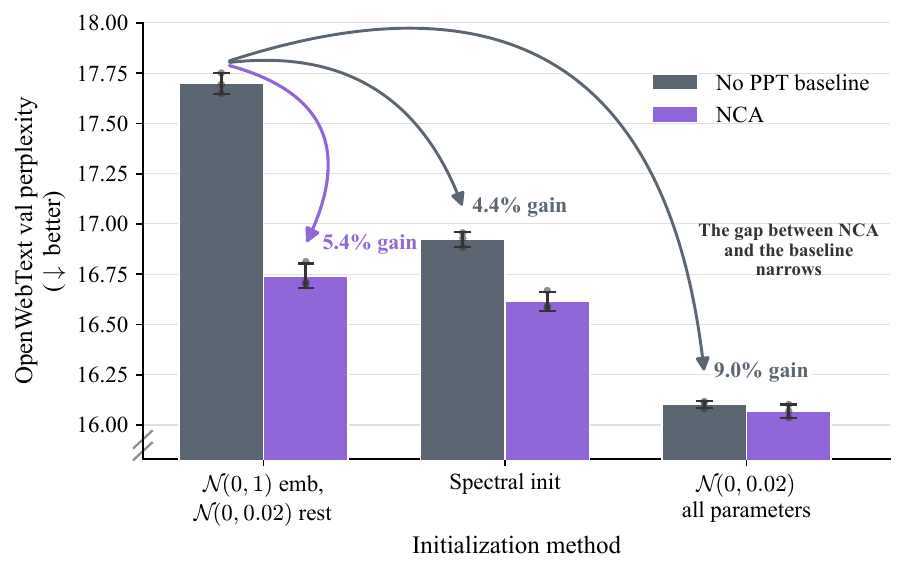}
\caption{\textbf{The NCA positive control under three baseline weight
initializations.} A 400M-parameter Llama model trained on 164M NCA tokens, and then ${\sim}9$B tokens of
OpenWebText, following \citet{han2026training}. Validation perplexity for the no-PPT baseline and NCA under three initializations, namely that of the original work ($\mathcal{N}(0,1)$ embeddings and $\mathcal{N}(0,0.02)$ elsewhere), $\mathcal{N}(0,0.02)$ embeddings and head, and the spectral-based initialization of the NanoChat version we use. Under each alternative initialization the baseline improves and the
gap between NCA and the baseline narrows.}
\label{fig:nca_init_v2}
\end{figure}

We compare three initializations, shown in Figure~\ref{fig:nca_init_v2}. The first is that of the original work: the embedding table is drawn from $\mathcal{N}(0,1)$, the transformer body from $\mathcal{N}(0,0.02)$, and the language-modeling head keeps PyTorch's default uniform initialization \citep{pytorch_linear, pytorch_tke}, whose standard deviation here is ${\approx}0.02$. The second draws both the token embedding table and the language-modeling head from
$\mathcal{N}(0,0.02)$. The third is the spectral-based initialization from the version of NanoChat we use:
linear layers from $\mathcal{N}(0,\sigma)$ with $\sigma = \min(1,
\sqrt{d_{\text{out}}/d_{\text{in}}})/\sqrt{d_{\text{in}}}$, the embedding table
from $\mathcal{N}(0,1)$, and the residual output projections and head zeroed. It
does not transfer to Llama exactly, as RMSNorm weights have no counterpart and
are left at ones.

Every run is initialized under the method being tested, including the PPT phase.
NCA therefore trains its PPT phase under that method, and at the transition the
new token embedding table and language-modeling head are drawn under the same
one, while the body carries over. As shown in Figure~\ref{fig:nca_init_v2}, the initialization that best suits the baseline is itself architecture-dependent: the spectral-based initialization is the default in the NanoChat version we use, but is not the best-performing of the three here. This provides further motivation to ablate initialization when evaluating PPT.

\section{FLOP Measurement}
\label{app:flop_measure}

To convert a FLOP budget into a training configuration, we measure the FLOPs of
one forward and backward pass (one optimization step) for the model using the
FLOP-counting utilities in PyTorch \citep{pytorchflopcounter}. We then divide the
total budget by the FLOPs per step to determine the number of optimization steps
and tokens consumed by each phase. Working backwards from a FLOP budget to a configuration is the
standard framing of compute-optimal PT \citep{hoffmann2022training},
and it is what allows a PPT method and its baseline to be compared on
equal terms for the compute-matched experiments. We measure FLOPs directly rather than matching token counts for compute matching because the synthetic corpora use much smaller vocabularies than natural language, so the language-modeling head, and with it the FLOPs per token, is smaller during PPT. Given its smaller vocabulary size, the PPT phase is therefore afforded extra steps. However, this difference shrinks as models grow, since the language-modeling head makes up a decreasing share of the total parameters. For the over-Chinchilla configuration, our largest model, we therefore simply allocate tokens directly, 0.5B to PPT and 49.5B to FineWeb against the baseline's 50B, so that every method sees the same total number of tokens (Appendix~\ref{app:overchinchilla}).

\section{Hyperparameters}
\label{app:hparams}

\subsection{How LLMTuna works}
\label{app:LLMTuna}

\texttt{LLMTuna} is a language-model-based hyperparameter optimizer that we release with this work. The approach is inspired by prior work that also used language models for hyperparameter optimization and autoresearch more broadly \citep{yang2023large, jiang2026autoresearch, karpathy2026autoresearch}. \texttt{LLMTuna} maintains an ordered history of configurations tried so far and their corresponding scores, which in our experiments are validation BPB. At each step, a language model uses this history together with any additional context provided, such as training code, notes on hyperparameter behavior, or constraints on the search, to propose the next configuration. This gives the model semantic context from the codebase that can inform which hyperparameters are likely to matter. The search space is specified using three parameter types, \texttt{Float}, \texttt{Int}, and \texttt{Choice}, each with a name, a natural-language description, and optional bounds and initial value.

Following the API design of Optuna \citep{akiba2019optuna}, the optimization loop consists of two calls. \texttt{suggest} provides the accumulated context to the model together with a tool whose schema is generated from the search space, so the proposal is returned as JSON in a fixed shape rather than as prose. \texttt{observe} then records the configuration and its resulting score, adding the trial to the history available to subsequent calls to \texttt{suggest}.

\subsection{Benchmark against Optuna}
\label{app:hpo_compare}

We compare LLMTuna against a Tree-structured Parzen Estimator (TPE) search using the Optuna library
\citep{akiba2019optuna} on one of our experiment configurations: NCA at 1\% and a total
budget of $1\times10^{18}$ FLOPs. Both searches minimize the same objective (validation set BPB) over
the same search space and initial hyperparameter configuration.

Optuna is run with the TPE sampler and a budget of 30
trials. The learning rate parameters are sampled in log space for a fair comparison, and
the constraint that the warmup and warmdown ratios sum to at most one is
enforced by conditional sampling, drawing the warmdown ratio from
$[0, 1 - \text{warmup}]$.
\texttt{LLMTuna} is run under the proposer, context and stopping rule given in
Appendix~\ref{app:search_config} (code repository in Appendix~\ref{app:code}). Results are shown in
Figure~\ref{fig:hpo_sample_efficiency}. These results suggest that \texttt{LLMTuna} can find good configurations with relatively few trials under the search constraints considered here.

\begin{figure}[h]
\centering
\includegraphics[width=0.7\linewidth]{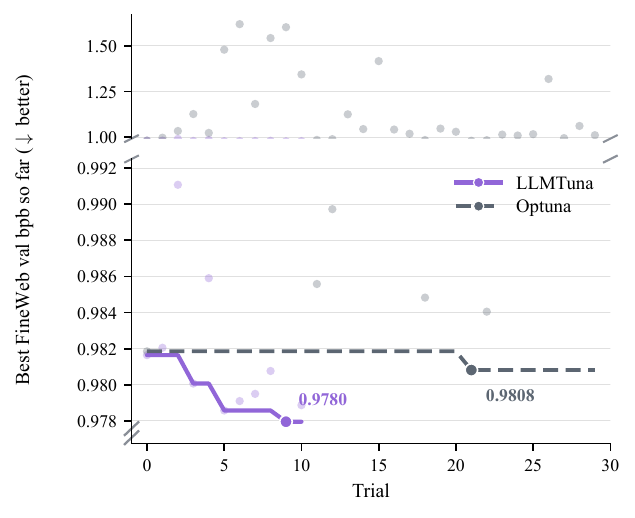}
\caption{\textbf{LLMTuna reaches a lower validation loss in fewer trials than a
TPE search.}
The continuous lines for each method show the best FineWeb validation BPB so far against trial index, while the faint dots are individual trials. LLMTuna reaches $0.9780$ at its tenth trial, while TPE reaches $0.9808$ at its twenty-second and does not improve
over the remaining eight. The axis is broken
above $0.992$, where the early exploratory trials of both searches sit.}
\label{fig:hpo_sample_efficiency}
\end{figure}

\subsection{Search configuration}
\label{app:search_config}

In the additive-compute setting, we run a separate \texttt{LLMTuna} search for each combination of PT natural-language token count, PPT token count, and the three weight-initialization schemes of Section~\ref{sec:setup}. In the data-constrained and data-unconstrained settings, a search is run for each combination of budget $C$, PPT fraction $\alpha$ and PPT corpus. Table~\ref{tab:hparams_searched} lists what each search tunes.

\begin{table}[h]
\centering
\small
\begin{tabular}{@{}llp{0.52\linewidth}@{}}
\toprule
\textbf{Setting} & \textbf{Recipe} & \textbf{Tuned hyperparameters} \\
\midrule
Additive & Unified  & Peak matrix learning rate, warmup fraction, warm\-down fraction, PT batch size \\
       & Separate & The above, plus PPT peak learning rate, PPT warmup fraction, PPT warm\-down fraction, PPT batch size\\
\midrule
Compute-matched & Unified  & Depth, peak matrix learning rate, warmup fraction, warm\-down fraction \\
              & Separate & The above, plus PPT peak learning rate, PPT warmup fraction, PPT warm\-down fraction, PPT batch size \\
\bottomrule
\end{tabular}
\caption{\textbf{Hyperparameters tuned in each setting.} \emph{Separate} is the
standard recipe of Section~\ref{sec:setup}: the PPT phase gets its own learning
rate, and the optimizer state is reset at the phase boundary.
\emph{Unified} covers the no-PPT baseline, the C4 control, and the unified optimizer code variant, which all run both
phases under a single schedule with no reset; for C4, restarting the optimizer
at the boundary hurt its final loss. In the additive setting the natural-language token count is fixed, so batch size sets the number of optimizer steps; we therefore tune it there, and fix it at 262,144 tokens per step in the compute-matched settings. The
PPT-phase optimizer is not searched; in the additive and data-constrained
settings we run one search under each of Muon and AdamW and report the
lower-BPB result.}
\label{tab:hparams_searched}
\end{table}

Every search is given a budget of at most 30 trials and stops early once five consecutive proposals fail to improve the best validation BPB by more than $10^{-3}$. Proposals come from \texttt{claude-sonnet-4-6} with a thinking budget of 300
tokens and a response cap of 4000, called through a tool interface that forces
it to return a value for every tuned hyperparameter. The proposer sees (i) a
system prompt naming only the hyperparameters that the study tunes, (ii) a
description of each, with its bounds or permitted options, (iii) a summary of
the training stack, generated once from ten source files and restricted to what
bears on the tuned hyperparameters, and (iv) every trial so far, with its full
configuration and validation BPB, in the order the trials were run. In the data-unconstrained
setting, each FLOP budget runs its own search, warm-started from the previous
budget's best configuration and carrying that search's history forward as
context, so the proposer can see how an optimum moved as the budget grew.

\subsection{Proposer prompt}
\label{app:proposer_prompt}

The system prompt for a study tuning all eight hyperparameters begins:

\begin{quote}\small\ttfamily
You are a hyperparameter optimizer minimizing val\_bpb (validation
bits-per-byte; lower is better).

You see all prior trials in the order they were run. Balance exploiting the current best
config with informative exploration; the trial budget is small, so every
proposal should either improve the best or teach you something specific.

Most of the result at a given scale comes from three knobs: depth, matrix\_lr,
and ppt\_lr. Each has its own optimum, and each optimum shifts with scale.
Getting them near their minimum for THIS scale matters more than refining the
other knobs, so make sure each has been searched widely enough that you would
notice if it were off.

\ldots
\end{quote}

The rest of the prompt covers step size and keeping proposals attributable; the
full prompts and hyperparameter definitions are in the released code.

\subsection{Hyperparameters found}
\label{app:hparams_found}

Tables~\ref{tab:hparams_constrained}--\ref{tab:hparams_interleave} list the
configuration of every cell the paper reports, one table per setting. Each table lists only the hyperparameters that differ between the
cells of its setting; the rest are shared by every run in the paper and are
stated here once. All models use a vocabulary of $65{,}536$ tokens (extended to hold the PPT tokens in the shared-vocabulary retention runs) and train at a sequence length of 2048 with a trapezoidal learning-rate schedule (linear warmdown to zero over the stated
fraction of the phase, no warmup unless noted), gradient clipping at 1.0 and
no weight decay. During natural-language pretraining Muon updates the matrix
parameters at the stated peak rate $\eta_{\mathrm{PT}}$ while the embedding
and unembedding tables train under AdamW at 0.3 and 0.004; during a PPT
phase all three groups share a single rate $\eta_{\mathrm{PPT}}$. Every value is
read from the logged configuration of the three seed replicas, which agree
on every field.

\begin{table}[h]
\centering
\footnotesize
\setlength{\tabcolsep}{4pt}
\input{tables/hparams_constrained}
\caption{\textbf{Data-constrained setting: reported configurations.} 100M
unique natural-language tokens at $1{\times}10^{18}$ FLOPs, the experiments of
panel~(1) of Figure~\ref{fig:combined_dose_scaling}, each at the lower-BPB
PPT optimizer. Every PPT run warms down over 0.4 of both its
PPT and natural-language phases, and the baseline over 0.2 of training.}
\label{tab:hparams_constrained}
\end{table}

\begin{table}[h]
\centering
\footnotesize
\setlength{\tabcolsep}{3pt}
\resizebox{\linewidth}{!}{\input{tables/hparams_scaling}}
\caption{\textbf{Data-unconstrained setting: reported configurations.} The
methods of Table~\ref{tab:scaling} at each budget. Two schedule details have no column: $k$-Dyck 1\% warms
its PPT phase up over the first 0.1 of the phase at every budget, and the
unified code 5\% method warms up over 0.02 of training at $3{\times}10^{18}$
and $9{\times}10^{18}$; every other experiment has no warmup.}
\label{tab:hparams_scaling}
\end{table}

\begin{table}[h]
\centering
\footnotesize
\setlength{\tabcolsep}{4pt}
\resizebox{\linewidth}{!}{\input{tables/hparams_additive}}
\caption{\textbf{Additive setting: reported configurations.} The twelve
points of Figure~\ref{fig:token_dose}: per natural-language (NL) budget, the two
baselines the figure draws and each curriculum at its lowest-loss
initialization, PPT token count and PPT optimizer. Every experiment settled on
$\eta_{\mathrm{PT}} = 0.02$ with a warmdown of 0.4, and the PPT phase runs
with no warmup.}
\label{tab:hparams_additive}
\end{table}

\begin{table}[h]
\centering
\footnotesize
\setlength{\tabcolsep}{4pt}
\input{tables/hparams_overchinchilla}
\caption{\textbf{Over-Chinchilla setting: configurations.} Depth 24 (881M
parameters) on 50B tokens, the methods of panel~(2) of
Figure~\ref{fig:combined_dose_scaling}. Hyperparameter search details are given in
Appendix~\ref{app:overchinchilla}.}
\label{tab:hparams_overchinchilla}
\end{table}

\begin{table}[h]
\centering
\footnotesize
\setlength{\tabcolsep}{4pt}
\input{tables/hparams_interleave}
\caption{\textbf{Retention experiment: configurations.} The runs of
Figure~\ref{fig:interleave} at $2.7{\times}10^{19}$ FLOPs. The PPT runs use
each method's 5\% configuration, extrapolated from the smaller budgets, with a
shared vocabulary. The interleaved runs train under one schedule, so they have
no PPT-phase entries, and all three share the code 5\% configuration so that
only the interleaved corpus differs.}
\label{tab:hparams_interleave}
\end{table}

\subsection{Over-Chinchilla configuration}
\label{app:overchinchilla}

The over-Chinchilla configuration's 50B tokens are ${\sim}12\times$ the
compute-optimal token count for its 881M-parameter model, based on the
${\approx}4.6$ tokens per parameter of the baseline's searched optima at
$1\times10^{18}$--$9\times10^{18}$ FLOPs. This is well below the $20$ token to parameter ratio reported by
\citet{hoffmann2022training}, as NanoChat's own fits also find
\citep{karpathy2026nanochatminiseries}. This is thanks to differences in data, model architecture, and optimizer. Nonetheless, even by the Chinchilla ratio, the run's
${\approx}57$ tokens per parameter is greater than the Chinchilla-optimal ratio.
We run the PPT methods at 1\% of total compute at this scale, which amounts to 0.5B tokens on PPT and 49.5B on FineWeb.

No \texttt{LLMTuna} search is run at this budget. Training uses no warmup and a
linear warmdown over the second half of training, and for the PPT runs the
warmdown fraction is set so that decay begins at the same 25B natural-language
token mark as in the baseline. We sweep only the peak learning rates, since the
optimum moves with the training horizon and this horizon is an order of
magnitude longer than any we searched. Every configuration is first run at
0.006, extrapolated from the baseline's searched optima, and then at three
further rates, 0.00748, 0.00563 and 0.00368, obtained by rescaling the peak rate
with the training horizon $T$ as $T^{p}$ for $p \in \{-0.2, -0.32, -0.5\}$, a
range informed by the power law between optimal learning rate and token
horizon reported by \citet{bjorck2025scaling}, with
each PPT run's PPT-phase rate rescaled by the same rule in its own horizon. Each
rate is run at a single seed, and the rate minimizing FineWeb validation BPB is
replicated across the remaining seeds: 0.006 for the baseline and $k$-Dyck,
0.00748 for NCA.

\section{Full results}
\label{app:extra}

This appendix collects the complete results for each setting in turn. For our reported
BLiMP results, we score BLiMP following the released evaluation code of
\citet{hu2025between}. Each of the 67,000 minimal pairs is scored by total
sentence log-probability, and a pair is counted correct only if the grammatical
sentence scores strictly higher than the ungrammatical one, so ties count as
errors. No beginning-of-sequence token is prepended, which means the first token of each sentence
is unscored. DCLM CORE is evaluated with the
NanoChat evaluation code \citep{nanochat}.
\label{app:blimp}

\clearpage
\subsection{Additive setting}
\label{app:additive_full}

\begin{figure}[h]
\centering
\includegraphics[width=\linewidth]{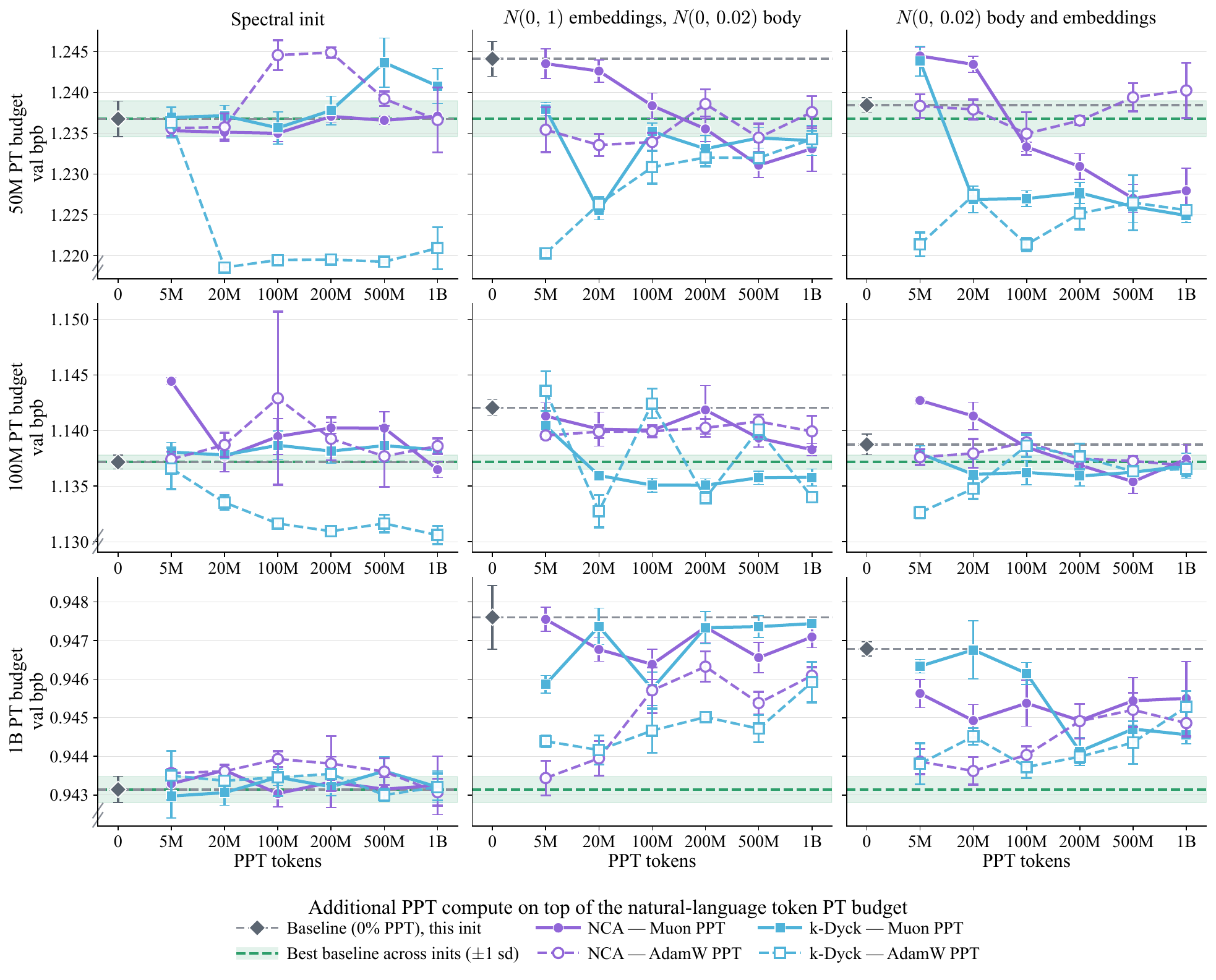}
\caption{\textbf{Full synthetic PPT token sweep behind
Figure~\ref{fig:token_dose}.} FineWeb
validation bits-per-byte against the amount of synthetic PPT added on
top of a fixed natural-language token budget. Rows are that budget, columns are the
initialization scheme: two as used in the prior literature that introduced these
curricula, and the spectral-based initialization. The model size here is 336M parameters. Within each panel the leftmost point is that configuration's own
baseline with no PPT, marked by the gray dashed line. The green dashed line and band mark the lowest-loss no-PPT baseline across the three initialization schemes, providing a reference for comparisons across initializations. Solid lines with
filled markers are Muon PPT and dashed lines with open markers are AdamW PPT. AdamW performs better than Muon for the PPT methods at the lower natural-language budgets; at 1B tokens, however, the two are indistinguishable under spectral-based initialization. Importantly, if one is to train with the alternative initializations, the PPT benefit would still exist even at this larger budget. }
\label{fig:token_dose_curves}
\end{figure}

Figure~\ref{fig:token_dose_curves} shows the full results behind Figure~\ref{fig:token_dose}, and Tables~\ref{tab:blimp_additive} and~\ref{tab:core_additive} report BLiMP and DCLM CORE accuracy for the same runs, with each curriculum at its lowest-loss PPT token count.

\begin{table}[h]
\centering
\small
\setlength{\tabcolsep}{4.5pt}
\input{tables/blimp_additive}
\caption{\textbf{BLiMP in the additive setting.}
BLiMP accuracy for the additive-compute experiments of
Figure~\ref{fig:token_dose}: for each natural-language budget and
initialization, the baseline and each curriculum at its best (lowest-loss) PPT
token count (Table~\ref{tab:hparams_additive} lists the counts). Higher is better, and bold marks values whose one-standard-deviation
interval lies entirely above that of the baseline with the same
initialization (none are bolded here as all perform similarly to the baseline).}
\label{tab:blimp_additive}
\end{table}

\begin{table}[h]
\centering
\small
\setlength{\tabcolsep}{4.5pt}
\input{tables/core_additive}
\caption{\textbf{DCLM CORE in the additive setting.}
Centered CORE accuracy for the same experiments as Table~\ref{tab:blimp_additive}:
for each natural-language budget and initialization, the baseline and each
curriculum at its best (lowest-loss) PPT token count
(Table~\ref{tab:hparams_additive} lists the counts). Higher is
better, and bold marks values whose one-standard-deviation interval lies
entirely above that of the baseline with the same initialization.}
\label{tab:core_additive}
\end{table}

\clearpage
\subsection{Data-unconstrained setting}
\label{app:scaling_unconstrained}

\begin{figure}[t]
\centering
\includegraphics[width=\linewidth]{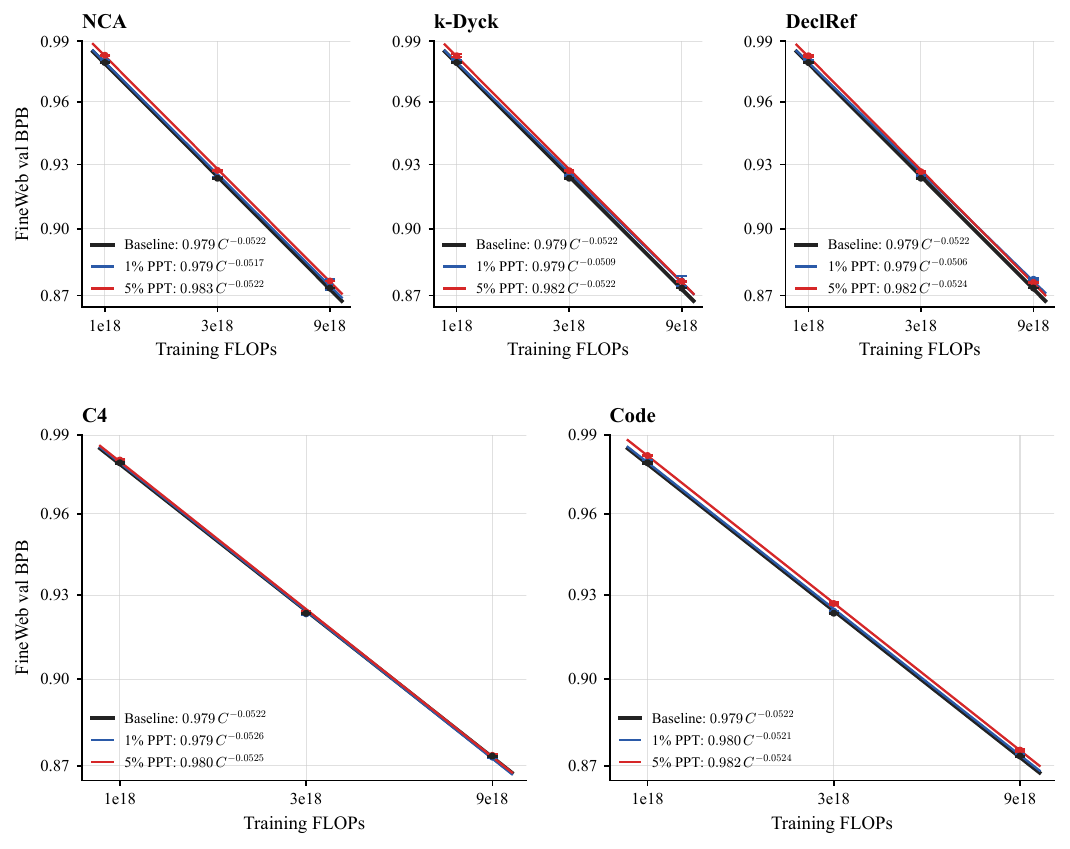}

\vspace{2pt}
{\footnotesize
\setlength{\tabcolsep}{4pt}
\renewcommand{\arraystretch}{0.9}
\input{tables/scaling_fits}}
\caption{\textbf{No PPT method scales better than the baseline.} Here we plot FineWeb validation
BPB against training compute for the data-unconstrained experiments. Each line is a separate power law, $\mathrm{BPB} = A\,C^{-\alpha}$
\citep{kaplan2020scalinglaws}, fitted by least squares on $\log \mathrm{BPB}$ against $\log C$
over that method's nine runs (three seeds per scale), with $C$ in units of $10^{18}$ FLOPs, so $A$ is
the fitted BPB at $10^{18}$ FLOPs. The table gives $A$ and $\alpha$ with their standard
errors. The baseline has the lowest $A$, and no method's exponent is significantly larger than the
baseline's (within standard error). Raising PPT from 1\% to 5\% raises $A$ for every method.
However, its effect on the exponent is unclear as at 5\% every exponent is within standard error of
the baseline's. Every synthetic PPT method remains above the baseline at every budget, while C4, independent of percentage, is the same as the baseline. Since we only have three budgets per method,
and three seeds, we do not fit an irreducible-loss term as in \citet{hoffmann2022training}.}
\label{fig:scaling_fits}
\end{figure}

Panel~(3) of Figure~\ref{fig:combined_dose_scaling} shows that, at each of the three budgets we
ran, spending compute on synthetic PPT does not beat spending it on natural language. A natural
follow-up is whether any method improves faster with compute, and so might overtake the
baseline at budgets beyond ours. Figure~\ref{fig:scaling_fits} addresses this by fitting a power
law to each method's FineWeb BPB across the same runs. The baseline starts lowest, and no PPT
method has a significantly larger scaling exponent, so we find no clear evidence that the gaps in
panel~(3) close with scale. Table~\ref{tab:scaling} gives the FineWeb BPB behind both figures. To
see how these results carry over to other evaluations, Tables~\ref{tab:scaling_c4}
and~\ref{tab:scaling_code} report BPB on the held-out C4 and GitHub-code splits for the same
checkpoints, and Tables~\ref{tab:blimp_scaling} and~\ref{tab:core_scaling} report BLiMP and DCLM
CORE accuracy.

\begin{table}[t]
\centering
\small
\setlength{\tabcolsep}{4.5pt}
\input{tables/scaling_bpb}
\caption{\textbf{Allocating a larger share of a fixed compute budget to synthetic
PPT generally increases loss.} FineWeb bits-per-byte. Hyperparameters were
searched independently for each method at every budget.
Code appears under both the separate and the unified optimizer recipe.
Bold marks values whose one-standard-deviation interval overlaps that of the column's lowest.}
\label{tab:scaling}
\end{table}

\begin{table}[h]
\centering
\small
\setlength{\tabcolsep}{4.5pt}
\input{tables/scaling_c4}
\caption{\textbf{The data-unconstrained comparison on the held-out C4 split.}
C4 bits-per-byte for the methods of Table~\ref{tab:scaling}, same checkpoints,
same protocol. Bold marks values
whose one-standard-deviation interval overlaps that of the column's lowest.}
\label{tab:scaling_c4}
\end{table}

\begin{table}[h]
\centering
\small
\setlength{\tabcolsep}{4.5pt}
\input{tables/scaling_code}
\caption{\textbf{The data-unconstrained comparison on the held-out GitHub-code
split.} %
GitHub-code bits-per-byte for the methods of Table~\ref{tab:scaling}, same
checkpoints, same protocol. Bold marks values whose one-standard-deviation interval
overlaps that of the column's lowest.}
\label{tab:scaling_code}
\end{table}

\begin{table}[h]
\centering
\small
\setlength{\tabcolsep}{4.5pt}
\input{tables/blimp_scaling}
\caption{\textbf{BLiMP across the data-unconstrained sweep.}
BLiMP accuracy for the data-unconstrained sweep; higher is better, and bold marks values whose one-standard-deviation
interval lies entirely above the baseline's at that budget.}
\label{tab:blimp_scaling}
\end{table}

\begin{table}[h]
\centering
\small
\setlength{\tabcolsep}{4.5pt}
\input{tables/core_scaling}
\caption{\textbf{DCLM CORE across the data-unconstrained sweep.}
Centered CORE accuracy; higher is better, and bold marks values whose one-standard-deviation interval lies entirely above the baseline's.}
\label{tab:core_scaling}
\end{table}

\subsection{Data-constrained setting}

\begin{figure}[h]
\centering
\includegraphics[width=0.72\linewidth]{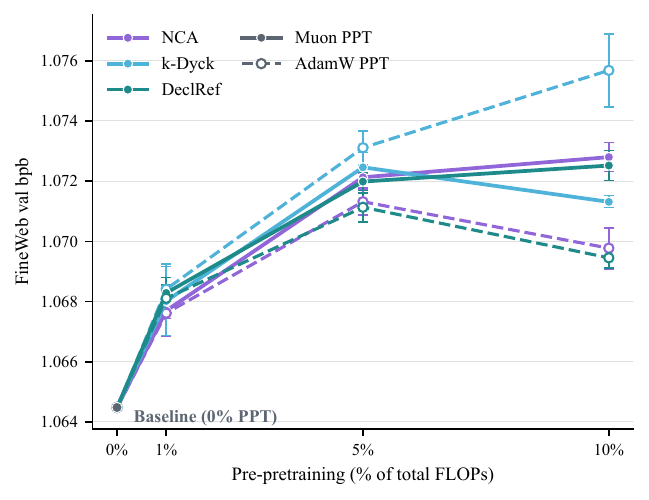}
\caption{\textbf{Data-constrained results separated by PPT-phase
optimizer.} The experiments behind panel~(1) of
Figure~\ref{fig:combined_dose_scaling}, with Muon PPT drawn as solid lines with
filled markers and AdamW PPT as dashed lines with open markers. Each experiment was
searched independently under the same search space and stopping rule, so the two
differ in the optimizer used during PPT and in the configuration each search found. Neither optimizer
is always better than the other: AdamW is lower for NCA and DeclRef at every PPT percentage, while $k$-Dyck is
lower under Muon. No experiment of either optimizer reaches the no-PPT baseline.}
\label{fig:constrained_dose_optimizers}
\end{figure}

Panel~(1) of Figure~\ref{fig:combined_dose_scaling} reports, for each curriculum
and PPT percentage, the best configuration found by the search. Because we run one search
under each PPT-phase optimizer and report the lower-BPB result, that panel
selects over the optimizer as well as over depth and learning rate. Figure~\ref{fig:constrained_dose_optimizers}
separates the two optimizers, and the benchmark scores for the same experiments follow.

\begin{table}[h]
\centering
\small
\setlength{\tabcolsep}{4.5pt}
\input{tables/constrained_bench}
\caption{\textbf{BLiMP and DCLM CORE in the data-constrained regime.}
100M unique natural-language tokens at a fixed $1{\times}10^{18}$-FLOP budget,
the experiments of panel~(1) of Figure~\ref{fig:combined_dose_scaling}, each at the
lower-BPB PPT optimizer. Higher is better, and bold marks values whose one-standard-deviation interval
lies entirely above the baseline's.}
\label{tab:blimp_constrained}
\end{table}

\clearpage

\subsection{Over-Chinchilla setting}

Benchmark scores for the 881M-parameter runs on 50B tokens.

\begin{table}[h]
\centering
\small
\setlength{\tabcolsep}{4.5pt}
\input{tables/overchinchilla_bench}
\caption{\textbf{Over-Chinchilla benchmark summary.} %
BLiMP and DCLM CORE for the over-Chinchilla regime: 881M parameters on 50B
tokens, NCA and $k$-Dyck at 1\%. Higher is better, and bold marks values whose one-standard-deviation interval lies entirely above the baseline's. No
method separates from the baseline on either benchmark.}
\label{tab:overchinchilla_bench}
\end{table}

\subsection{Downstream benchmarks}
\label{app:benchmarks}

The DCLM CORE suite from \citet{li2025datacomplmsearchgenerationtraining} aggregates 22 tasks: AGIEval
LSAT-AR, ARC-Easy, ARC-Challenge, six BIG-bench tasks (QA Wikidata, Dyck
languages, Operators, Repeat Copy Logic, CS Algorithms, and Language
Identification), BoolQ, CommonsenseQA, COPA, CoQA, HellaSwag (zero-shot and
10-shot), Jeopardy, LAMBADA, OpenBookQA, PIQA, SQuAD, the Winograd Schema
Challenge, and Winogrande. Please refer to the original paper for more details on each benchmark.

\begin{figure}[p]
\centering
\includegraphics[width=\linewidth]{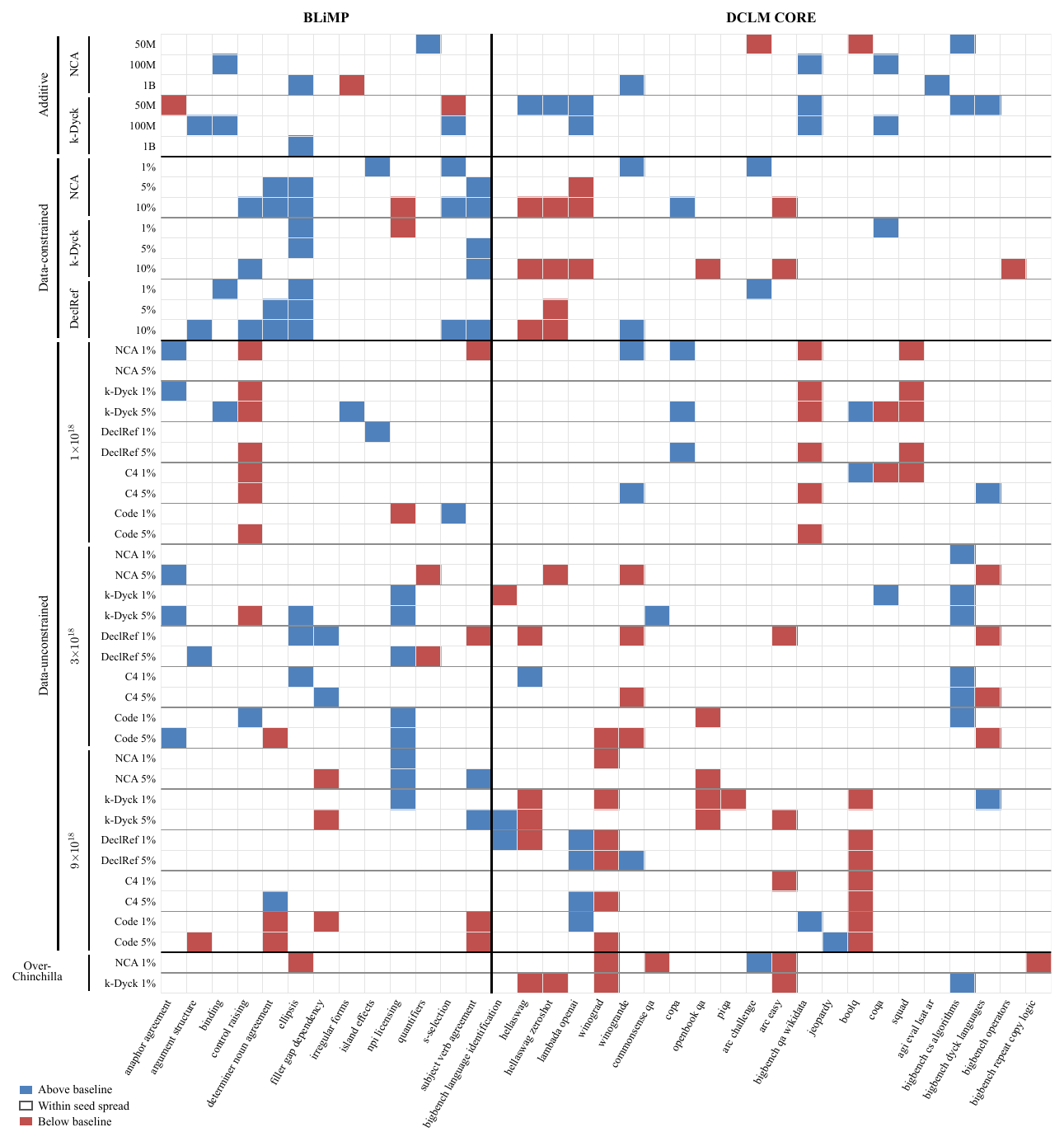}
\caption{\textbf{Breakdown of downstream performance by benchmark sub-tasks.} Each row is a PPT model from Figures~\ref{fig:combined_dose_scaling} and~\ref{fig:token_dose}, and each column is one of the 13 BLiMP or 22 DCLM CORE tasks. Blue cells are significantly above the baseline, red cells significantly below, and white cells on par. Most cells are white, so on most sub-scores the PPT models perform on par with their baseline. The colored cells show that, even so, the PPT method changes how the model performs on some sub-tasks of these benchmarks.}
\label{fig:subscore_direction}
\end{figure}

\section{Code release}
\label{app:code}

The training and evaluation code is available at \href{https://github.com/alexkstern/rethinking-ppt}{\texttt{rethinking-ppt}}, and \texttt{LLMTuna} at \href{https://github.com/alexkstern/llmtuna-hpo}{\texttt{llmtuna-hpo}}.
\end{document}

%% file: math_commands.tex
\usepackage{amsmath,amsfonts,bm}

\def\eqref#1{equation~\ref{#1}}

\def\1{\bm{1}}

\DeclareMathAlphabet{\mathsfit}{\encodingdefault}{\sfdefault}{m}{sl}
\SetMathAlphabet{\mathsfit}{bold}{\encodingdefault}{\sfdefault}{bx}{n}

%% file: tables/aggregate_wins.tex
\begin{tabular}{l l l c c}
\toprule
 & Setting & PPT method & Baseline (0\% PPT) & PPT score \\
\midrule
\begin{tabular}[t]{@{}l@{}}BLiMP\\($\uparrow$ better)\end{tabular} & Data-constrained (100M NL, $1{\times}10^{18}$) & DeclRef 10\% & 0.721 $\pm$ 0.004 & 0.731 $\pm$ 0.005 \\
\midrule
\multirow{5}{*}{\shortstack[l]{DCLM CORE\\($\uparrow$ better)}} & Additive (50M NL) & $k$-Dyck & 0.024 $\pm$ 0.004 & 0.032 $\pm$ 0.003 \\
\cmidrule(lr){2-5}
 & \multirow{4}{*}{Data-unconstrained ($1{\times}10^{18}$)} & C4 1\% & \multirow{4}{*}{0.088 $\pm$ 0.006} & 0.106 $\pm$ 0.004 \\
 &  & Code 1\% &  & 0.104 $\pm$ 0.004 \\
 &  & DeclRef 1\% &  & 0.097 $\pm$ 0.002 \\
 &  & DeclRef 5\% &  & 0.108 $\pm$ 0.011 \\
\bottomrule
\end{tabular}

%% file: tables/hparams_constrained.tex
\begin{tabular}{l c c c c c}
\toprule
Method & $d$ & $\eta_{\mathrm{PT}}$ & PPT optimizer & $\eta_{\mathrm{PPT}}$ & PPT batch \\
\midrule
Baseline (0\% PPT) & 20 & 3e-2 & -- & -- & -- \\
NCA 1\% & 20 & 3e-2 & AdamW & 3e-5 & 512k \\
NCA 5\% & 20 & 3e-2 & AdamW & 3e-5 & 512k \\
NCA 10\% & 22 & 2e-2 & AdamW & 3e-6 & 1M \\
$k$-Dyck 1\% & 20 & 3e-2 & Muon & 8e-6 & 512k \\
$k$-Dyck 5\% & 20 & 3e-2 & Muon & 2e-2 & 512k \\
$k$-Dyck 10\% & 22 & 3e-2 & Muon & 1e-3 & 1M \\
DeclRef 1\% & 20 & 3e-2 & AdamW & 3e-5 & 256k \\
DeclRef 5\% & 20 & 3e-2 & AdamW & 3e-5 & 512k \\
DeclRef 10\% & 21 & 2e-2 & AdamW & 1e-6 & 128k \\
\bottomrule
\end{tabular}

%% file: tables/hparams_scaling.tex
\begin{tabular}{l ccc ccc ccc ccc ccc ccc}
\toprule
& \multicolumn{3}{c}{$d$} & \multicolumn{3}{c}{$\eta_{\mathrm{PT}}$} & \multicolumn{3}{c}{$\eta_{\mathrm{PPT}}$} & \multicolumn{3}{c}{PT warmdown} & \multicolumn{3}{c}{PPT warmdown} & \multicolumn{3}{c}{PPT batch} \\
\cmidrule(lr){2-4} \cmidrule(lr){5-7} \cmidrule(lr){8-10} \cmidrule(lr){11-13} \cmidrule(lr){14-16} \cmidrule(lr){17-19}
Method & $1{\times}10^{18}$ & $3{\times}10^{18}$ & $9{\times}10^{18}$ & $1{\times}10^{18}$ & $3{\times}10^{18}$ & $9{\times}10^{18}$ & $1{\times}10^{18}$ & $3{\times}10^{18}$ & $9{\times}10^{18}$ & $1{\times}10^{18}$ & $3{\times}10^{18}$ & $9{\times}10^{18}$ & $1{\times}10^{18}$ & $3{\times}10^{18}$ & $9{\times}10^{18}$ & $1{\times}10^{18}$ & $3{\times}10^{18}$ & $9{\times}10^{18}$ \\
\midrule
Baseline (0\% PPT) & 12 & 16 & 20 & 3e-2 & 2e-2 & 1.5e-2 & -- & -- & -- & 0.4 & 0.4 & 0.5 & -- & -- & -- & -- & -- & -- \\
\midrule
C4 1\% & 11 & 15 & 19 & 3e-2 & 1.5e-2 & 1.5e-2 & -- & -- & -- & 0.4 & 0.4 & 0.4 & -- & -- & -- & -- & -- & -- \\
C4 5\% & 12 & 15 & 19 & 3e-2 & 1.5e-2 & 1.5e-2 & -- & -- & -- & 0.4 & 0.4 & 0.4 & -- & -- & -- & -- & -- & -- \\
\midrule
NCA 1\% & 11 & 15 & 19 & 2e-2 & 2e-2 & 2e-2 & 6e-3 & 2e-2 & 1.2e-2 & 0.4 & 0.4 & 0.4 & 0.4 & 0.4 & 0.4 & 1M & 1M & 1M \\
NCA 5\% & 12 & 16 & 20 & 3e-2 & 3e-2 & 3e-2 & 2e-3 & 6e-5 & 6e-5 & 0.4 & 0.4 & 0.4 & 0.4 & 0.4 & 0.4 & 256k & 256k & 256k \\
$k$-Dyck 1\% & 11 & 16 & 20 & 3e-2 & 3e-2 & 4e-2 & 3e-3 & 1e-2 & 1e-2 & 0.4 & 0.4 & 0.4 & 0.4 & 0.4 & 0.4 & 1M & 1M & 1M \\
$k$-Dyck 5\% & 12 & 16 & 18 & 3e-2 & 3e-2 & 3e-2 & 3e-5 & 3e-5 & 3e-5 & 0.4 & 0.4 & 0.4 & 0.4 & 0.4 & 0.4 & 512k & 512k & 512k \\
DeclRef 1\% & 12 & 16 & 18 & 3e-2 & 1.5e-2 & 1.5e-2 & 1.2e-2 & 1.2e-2 & 1.2e-2 & 0.4 & 0.4 & 0.4 & 0.4 & 0.4 & 0.4 & 256k & 256k & 256k \\
DeclRef 5\% & 12 & 16 & 20 & 2e-2 & 1.5e-2 & 1.5e-2 & 2e-3 & 1e-3 & 1e-3 & 0.4 & 0.4 & 0.4 & 0.4 & 0.4 & 0.4 & 256k & 256k & 256k \\
\midrule
Code 1\% (unified optimizer) & 11 & 15 & 18 & 3e-2 & 3e-2 & 1.5e-2 & -- & -- & -- & 0.4 & 0.5 & 0.5 & -- & -- & -- & -- & -- & -- \\
Code 5\% (unified optimizer) & 12 & 16 & 20 & 1.5e-2 & 1.5e-2 & 1.5e-2 & -- & -- & -- & 0.4 & 0.5 & 0.5 & -- & -- & -- & -- & -- & -- \\
Code 1\% (separate PPT optimizer) & 11 & 15 & 18 & 3e-2 & 3e-2 & 2e-2 & 1e-4 & 3e-4 & 3e-4 & 0.4 & 0.4 & 0.4 & 0.4 & 0.7 & 0.7 & 1M & 1M & 1M \\
Code 5\% (separate PPT optimizer) & 12 & 14 & 18 & 3e-2 & 3e-2 & 2e-2 & 1e-5 & 1e-5 & 1e-5 & 0.4 & 0.4 & 0.4 & 0.4 & 0.8 & 0.8 & 256k & 256k & 512k \\
\bottomrule
\end{tabular}

%% file: tables/hparams_additive.tex
\begin{tabular}{l l l c c c c c c}
\toprule
NL tokens & Method & Init & PPT tokens & PT batch & PPT optimizer & $\eta_{\mathrm{PPT}}$ & PPT warmdown & PPT batch \\
\midrule
50M & Baseline & $\mathcal{N}(0,1)$ emb. & -- & 64k & -- & -- & -- & -- \\
50M & Baseline & spectral & -- & 64k & -- & -- & -- & -- \\
50M & NCA & $\mathcal{N}(0,0.02)$ & 500M & 64k & Muon & 4e-3 & 1 & 256k \\
50M & $k$-Dyck & spectral & 20M & 64k & AdamW & 1.2e-4 & 1 & 64k \\
\midrule
100M & Baseline & $\mathcal{N}(0,1)$ emb. & -- & 64k & -- & -- & -- & -- \\
100M & Baseline & spectral & -- & 64k & -- & -- & -- & -- \\
100M & NCA & $\mathcal{N}(0,0.02)$ & 500M & 64k & Muon & 6e-3 & 1 & 512k \\
100M & $k$-Dyck & spectral & 1B & 64k & AdamW & 5e-5 & 1 & 512k \\
\midrule
1B & Baseline & $\mathcal{N}(0,1)$ emb. & -- & 256k & -- & -- & -- & -- \\
1B & Baseline & spectral & -- & 256k & -- & -- & -- & -- \\
1B & NCA & spectral & 100M & 256k & Muon & 6e-6 & 1 & 256k \\
1B & $k$-Dyck & spectral & 5M & 256k & Muon & 6e-6 & 0.4 & 256k \\
\bottomrule
\end{tabular}

%% file: tables/hparams_overchinchilla.tex
\begin{tabular}{l c c c c c c}
\toprule
Method & $\eta_{\mathrm{PT}}$ & PT warmdown & $\eta_{\mathrm{PPT}}$ & PPT warmup & PPT warmdown & PPT batch \\
\midrule
Baseline (0\% PPT) & 6e-3 & 0.5 & -- & -- & -- & -- \\
NCA 1\% & 7.5e-3 & 0.495 & 1.3e-3 & 0 & 0.6 & 1M \\
$k$-Dyck 1\% & 6e-3 & 0.495 & 1e-2 & 0.1 & 0.4 & 1M \\
\bottomrule
\end{tabular}

%% file: tables/hparams_interleave.tex
\begin{tabular}{l c c c c c c}
\toprule
Run & $d$ & $\eta_{\mathrm{PT}}$ & PT warmdown & $\eta_{\mathrm{PPT}}$ & PPT warmdown & PPT batch \\
\midrule
NCA 5\%, interleaved & 24 & 1.5e-2 & 0.5 & -- & -- & -- \\
$k$-Dyck 5\%, interleaved & 24 & 1.5e-2 & 0.5 & -- & -- & -- \\
Code 5\%, interleaved & 24 & 1.5e-2 & 0.5 & -- & -- & -- \\
NCA 5\%, PPT phase & 24 & 3e-2 & 0.4 & 6e-5 & 0.4 & 256k \\
$k$-Dyck 5\%, PPT phase & 21 & 3e-2 & 0.4 & 3e-5 & 0.4 & 512k \\
Code 5\%, PPT phase & 24 & 1.5e-2 & 0.5 & -- & -- & -- \\
\bottomrule
\end{tabular}

%% file: tables/blimp_additive.tex
\begin{tabular}{ll ccc}
\toprule
Initialization & Method & 50M & 100M & 1B \\
\midrule
Spectral init & Baseline & 0.691 $\pm$ 0.005 & 0.719 $\pm$ 0.001 & 0.774 $\pm$ 0.005 \\
 & NCA & 0.691 $\pm$ 0.002 & 0.719 $\pm$ 0.008 & 0.769 $\pm$ 0.007 \\
 & $k$-Dyck & 0.688 $\pm$ 0.006 & 0.723 $\pm$ 0.007 & 0.778 $\pm$ 0.007 \\
\midrule
$N(0,1)$ embeddings & Baseline & 0.687 $\pm$ 0.003 & 0.717 $\pm$ 0.004 & 0.779 $\pm$ 0.003 \\
 & NCA & 0.683 $\pm$ 0.003 & 0.717 $\pm$ 0.007 & 0.776 $\pm$ 0.002 \\
 & $k$-Dyck & 0.695 $\pm$ 0.007 & 0.718 $\pm$ 0.011 & 0.775 $\pm$ 0.001 \\
\midrule
$N(0,0.02)$ all & Baseline & 0.695 $\pm$ 0.005 & 0.720 $\pm$ 0.006 & 0.779 $\pm$ 0.005 \\
 & NCA & 0.691 $\pm$ 0.008 & 0.724 $\pm$ 0.006 & 0.774 $\pm$ 0.004 \\
 & $k$-Dyck & 0.690 $\pm$ 0.005 & 0.722 $\pm$ 0.011 & 0.783 $\pm$ 0.005 \\
\bottomrule
\end{tabular}

%% file: tables/core_additive.tex
\begin{tabular}{ll ccc}
\toprule
Initialization & Method & 50M & 100M & 1B \\
\midrule
Spectral init & Baseline & 0.024 $\pm$ 0.004 & 0.042 $\pm$ 0.011 & 0.110 $\pm$ 0.009 \\
 & NCA & \textbf{0.032 $\pm$ 0.001} & 0.046 $\pm$ 0.006 & 0.120 $\pm$ 0.002 \\
 & $k$-Dyck & \textbf{0.032 $\pm$ 0.003} & 0.046 $\pm$ 0.012 & 0.116 $\pm$ 0.003 \\
\midrule
$N(0,1)$ embeddings & Baseline & 0.030 $\pm$ 0.002 & 0.040 $\pm$ 0.007 & 0.107 $\pm$ 0.010 \\
 & NCA & 0.035 $\pm$ 0.010 & 0.045 $\pm$ 0.009 & 0.117 $\pm$ 0.014 \\
 & $k$-Dyck & 0.030 $\pm$ 0.005 & 0.044 $\pm$ 0.006 & 0.115 $\pm$ 0.015 \\
\midrule
$N(0,0.02)$ all & Baseline & 0.031 $\pm$ 0.007 & 0.049 $\pm$ 0.008 & 0.121 $\pm$ 0.003 \\
 & NCA & 0.027 $\pm$ 0.006 & 0.047 $\pm$ 0.012 & 0.116 $\pm$ 0.006 \\
 & $k$-Dyck & 0.037 $\pm$ 0.012 & 0.044 $\pm$ 0.004 & 0.117 $\pm$ 0.012 \\
\bottomrule
\end{tabular}

%% file: tables/scaling_fits.tex
\begin{tabular}{l cc cc}
\toprule
 & \multicolumn{2}{c}{1\% PPT} & \multicolumn{2}{c}{5\% PPT} \\
\cmidrule(lr){2-3} \cmidrule(lr){4-5}
Method & $A$ & $\alpha$ & $A$ & $\alpha$ \\
\midrule
NCA & $0.9795 \pm 0.0010$ & $0.0517 \pm 0.0008$ & $0.9827 \pm 0.0003$ & $0.0522 \pm 0.0002$ \\
$k$-Dyck & $0.9794 \pm 0.0009$ & $0.0509 \pm 0.0007$ & $0.9824 \pm 0.0004$ & $0.0522 \pm 0.0003$ \\
DeclRef & $0.9794 \pm 0.0009$ & $0.0506 \pm 0.0006$ & $0.9823 \pm 0.0004$ & $0.0524 \pm 0.0003$ \\
C4 & $0.9793 \pm 0.0006$ & $0.0526 \pm 0.0004$ & $0.9799 \pm 0.0005$ & $0.0525 \pm 0.0004$ \\
Code & $0.9795 \pm 0.0004$ & $0.0521 \pm 0.0003$ & $0.9821 \pm 0.0002$ & $0.0524 \pm 0.0002$ \\
\midrule
Baseline (0\% PPT) & \multicolumn{4}{c}{$A = \mathbf{0.9789} \pm 0.0004$, \quad $\alpha = 0.0522 \pm 0.0003$} \\
\bottomrule
\end{tabular}

%% file: tables/scaling_bpb.tex
\begin{tabular}{l ccc}
\toprule
Method & $1{\times}10^{18}$ & $3{\times}10^{18}$ & $9{\times}10^{18}$ \\
\midrule
Baseline (0\% PPT) & $\mathbf{0.9794}_{\pm 0.0003}$ & $\mathbf{0.9234}_{\pm 0.0002}$ & $\mathbf{0.8733}_{\pm 0.0002}$ \\
\midrule
C4 1\% & $\mathbf{0.9799}_{\pm 0.0006}$ & $\mathbf{0.9232}_{\pm 0.0005}$ & $\mathbf{0.8729}_{\pm 0.0004}$ \\
C4 5\% & $0.9804_{\pm 0.0004}$ & $\mathbf{0.9238}_{\pm 0.0003}$ & $\mathbf{0.8735}_{\pm 0.0005}$ \\
\midrule
NCA 1\% & $0.9803_{\pm 0.0005}$ & $\mathbf{0.9239}_{\pm 0.0001}$ & $\mathbf{0.8751}_{\pm 0.0024}$ \\
NCA 5\% & $0.9830_{\pm 0.0002}$ & $0.9272_{\pm 0.0002}$ & $0.8765_{\pm 0.0002}$ \\
$k$-Dyck 1\% & $0.9802_{\pm 0.0002}$ & $0.9248_{\pm 0.0003}$ & $0.8765_{\pm 0.0021}$ \\
$k$-Dyck 5\% & $0.9828_{\pm 0.0007}$ & $0.9270_{\pm 0.0001}$ & $0.8763_{\pm 0.0003}$ \\
DeclRef 1\% & $0.9804_{\pm 0.0004}$ & $0.9245_{\pm 0.0005}$ & $0.8772_{\pm 0.0004}$ \\
DeclRef 5\% & $0.9827_{\pm 0.0005}$ & $0.9265_{\pm 0.0001}$ & $0.8758_{\pm 0.0003}$ \\
\midrule
Code 1\% (unified optimizer) & $0.9831_{\pm 0.0003}$ & $0.9266_{\pm 0.0002}$ & $0.8790_{\pm 0.0056}$ \\
Code 5\% (unified optimizer) & $0.9869_{\pm 0.0003}$ & $0.9305_{\pm 0.0002}$ & $0.8817_{\pm 0.0016}$ \\
Code 1\% (separate PPT optimizer) & $\mathbf{0.9800}_{\pm 0.0004}$ & $0.9242_{\pm 0.0002}$ & $0.8740_{\pm 0.0001}$ \\
Code 5\% (separate PPT optimizer) & $0.9822_{\pm 0.0003}$ & $0.9270_{\pm 0.0004}$ & $0.8753_{\pm 0.0005}$ \\
\bottomrule
\end{tabular}

%% file: tables/scaling_c4.tex
\begin{tabular}{l ccc}
\toprule
Method & $1{\times}10^{18}$ & $3{\times}10^{18}$ & $9{\times}10^{18}$ \\
\midrule
Baseline (0\% PPT) & $\mathbf{1.1461}_{\pm 0.0002}$ & $1.0900_{\pm 0.0002}$ & $1.0409_{\pm 0.0004}$ \\
\midrule
C4 1\% & $\mathbf{1.1465}_{\pm 0.0007}$ & $1.0898_{\pm 0.0007}$ & $1.0399_{\pm 0.0005}$ \\
C4 5\% & $\mathbf{1.1456}_{\pm 0.0004}$ & $\mathbf{1.0885}_{\pm 0.0004}$ & $\mathbf{1.0382}_{\pm 0.0008}$ \\
\midrule
NCA 1\% & $1.1471_{\pm 0.0006}$ & $1.0905_{\pm 0.0003}$ & $1.0424_{\pm 0.0027}$ \\
NCA 5\% & $1.1497_{\pm 0.0003}$ & $1.0937_{\pm 0.0003}$ & $1.0433_{\pm 0.0002}$ \\
$k$-Dyck 1\% & $1.1470_{\pm 0.0002}$ & $1.0914_{\pm 0.0003}$ & $1.0436_{\pm 0.0021}$ \\
$k$-Dyck 5\% & $1.1493_{\pm 0.0006}$ & $1.0935_{\pm 0.0001}$ & $1.0443_{\pm 0.0005}$ \\
DeclRef 1\% & $1.1472_{\pm 0.0006}$ & $1.0912_{\pm 0.0006}$ & $1.0446_{\pm 0.0005}$ \\
DeclRef 5\% & $1.1493_{\pm 0.0007}$ & $1.0931_{\pm 0.0001}$ & $1.0428_{\pm 0.0005}$ \\
\midrule
Code 1\% (unified optimizer) & $1.1500_{\pm 0.0005}$ & $1.0935_{\pm 0.0003}$ & $1.0468_{\pm 0.0054}$ \\
Code 5\% (unified optimizer) & $1.1537_{\pm 0.0003}$ & $1.0973_{\pm 0.0002}$ & $1.0487_{\pm 0.0017}$ \\
Code 1\% (separate PPT optimizer) & $1.1468_{\pm 0.0004}$ & $1.0907_{\pm 0.0002}$ & $1.0420_{\pm 0.0005}$ \\
Code 5\% (separate PPT optimizer) & $1.1488_{\pm 0.0004}$ & $1.0938_{\pm 0.0002}$ & $1.0432_{\pm 0.0006}$ \\
\bottomrule
\end{tabular}

%% file: tables/scaling_code.tex
\begin{tabular}{l ccc}
\toprule
Method & $1{\times}10^{18}$ & $3{\times}10^{18}$ & $9{\times}10^{18}$ \\
\midrule
Baseline (0\% PPT) & $\mathbf{1.9435}_{\pm 0.0260}$ & $\mathbf{1.7078}_{\pm 0.0785}$ & $\mathbf{1.6702}_{\pm 0.0347}$ \\
\midrule
C4 1\% & $\mathbf{1.9738}_{\pm 0.0498}$ & $\mathbf{1.7890}_{\pm 0.0528}$ & $\mathbf{1.6590}_{\pm 0.0286}$ \\
C4 5\% & $\mathbf{1.9012}_{\pm 0.0132}$ & $\mathbf{1.7914}_{\pm 0.0732}$ & $\mathbf{1.6832}_{\pm 0.0217}$ \\
\midrule
NCA 1\% & $\mathbf{1.9356}_{\pm 0.0883}$ & $\mathbf{1.7793}_{\pm 0.0469}$ & $\mathbf{1.7353}_{\pm 0.0295}$ \\
NCA 5\% & $\mathbf{1.9640}_{\pm 0.0603}$ & $\mathbf{1.8043}_{\pm 0.0862}$ & $\mathbf{1.7272}_{\pm 0.0227}$ \\
$k$-Dyck 1\% & $\mathbf{1.9867}_{\pm 0.0929}$ & $\mathbf{1.7675}_{\pm 0.0580}$ & $\mathbf{1.8248}_{\pm 0.1379}$ \\
$k$-Dyck 5\% & $1.9536_{\pm 0.0281}$ & $\mathbf{1.8378}_{\pm 0.1098}$ & $\mathbf{1.8205}_{\pm 0.1192}$ \\
DeclRef 1\% & $1.9933_{\pm 0.0670}$ & $\mathbf{1.7080}_{\pm 0.1051}$ & $\mathbf{1.7438}_{\pm 0.0528}$ \\
DeclRef 5\% & $\mathbf{1.9250}_{\pm 0.0147}$ & $\mathbf{1.7385}_{\pm 0.0628}$ & $\mathbf{1.7206}_{\pm 0.0545}$ \\
\midrule
Code 1\% (unified optimizer) & $\mathbf{1.9601}_{\pm 0.0834}$ & $\mathbf{1.8063}_{\pm 0.1156}$ & $\mathbf{1.7206}_{\pm 0.0346}$ \\
Code 5\% (unified optimizer) & $\mathbf{1.8892}_{\pm 0.0361}$ & $\mathbf{1.7491}_{\pm 0.0304}$ & $\mathbf{1.6568}_{\pm 0.0585}$ \\
Code 1\% (separate PPT optimizer) & $\mathbf{1.9697}_{\pm 0.0891}$ & $\mathbf{1.7372}_{\pm 0.0510}$ & $\mathbf{1.6882}_{\pm 0.0374}$ \\
Code 5\% (separate PPT optimizer) & $\mathbf{1.9426}_{\pm 0.0390}$ & $1.7975_{\pm 0.0099}$ & $\mathbf{1.6924}_{\pm 0.0661}$ \\
\bottomrule
\end{tabular}

%% file: tables/blimp_scaling.tex
\begin{tabular}{l ccc}
\toprule
Method & $1{\times}10^{18}$ & $3{\times}10^{18}$ & $9{\times}10^{18}$ \\
\midrule
Baseline (0\% PPT) & 0.761 $\pm$ 0.004 & 0.780 $\pm$ 0.003 & 0.794 $\pm$ 0.004 \\
\midrule
C4 1\% & 0.756 $\pm$ 0.007 & 0.779 $\pm$ 0.004 & 0.791 $\pm$ 0.009 \\
C4 5\% & 0.757 $\pm$ 0.009 & 0.780 $\pm$ 0.005 & 0.794 $\pm$ 0.008 \\
\midrule
NCA 1\% & 0.753 $\pm$ 0.006 & 0.781 $\pm$ 0.007 & 0.792 $\pm$ 0.002 \\
NCA 5\% & 0.759 $\pm$ 0.006 & 0.773 $\pm$ 0.010 & 0.796 $\pm$ 0.002 \\
$k$-Dyck 1\% & 0.755 $\pm$ 0.007 & 0.782 $\pm$ 0.006 & 0.798 $\pm$ 0.002 \\
$k$-Dyck 5\% & 0.762 $\pm$ 0.006 & 0.783 $\pm$ 0.001 & 0.797 $\pm$ 0.002 \\
DeclRef 1\% & 0.758 $\pm$ 0.013 & 0.779 $\pm$ 0.010 & 0.791 $\pm$ 0.009 \\
DeclRef 5\% & 0.754 $\pm$ 0.002 & 0.778 $\pm$ 0.003 & 0.797 $\pm$ 0.008 \\
\midrule
Code 1\% (unified optimizer) & 0.753 $\pm$ 0.004 & 0.782 $\pm$ 0.002 & 0.789 $\pm$ 0.007 \\
Code 5\% (unified optimizer) & 0.762 $\pm$ 0.009 & 0.784 $\pm$ 0.004 & 0.793 $\pm$ 0.003 \\
Code 1\% (separate PPT optimizer) & 0.758 $\pm$ 0.007 & 0.783 $\pm$ 0.007 & 0.796 $\pm$ 0.006 \\
Code 5\% (separate PPT optimizer) & 0.755 $\pm$ 0.009 & 0.780 $\pm$ 0.005 & 0.789 $\pm$ 0.004 \\
\bottomrule
\end{tabular}

%% file: tables/core_scaling.tex
\begin{tabular}{l ccc}
\toprule
Method & $1{\times}10^{18}$ & $3{\times}10^{18}$ & $9{\times}10^{18}$ \\
\midrule
Baseline (0\% PPT) & 0.088 $\pm$ 0.006 & 0.121 $\pm$ 0.015 & 0.165 $\pm$ 0.008 \\
\midrule
C4 1\% & \textbf{0.106 $\pm$ 0.004} & 0.131 $\pm$ 0.004 & 0.156 $\pm$ 0.004 \\
C4 5\% & 0.101 $\pm$ 0.007 & 0.137 $\pm$ 0.005 & 0.154 $\pm$ 0.007 \\
\midrule
NCA 1\% & 0.092 $\pm$ 0.005 & 0.127 $\pm$ 0.011 & 0.162 $\pm$ 0.007 \\
NCA 5\% & 0.082 $\pm$ 0.003 & 0.133 $\pm$ 0.004 & 0.155 $\pm$ 0.012 \\
$k$-Dyck 1\% & 0.096 $\pm$ 0.012 & 0.135 $\pm$ 0.010 & 0.153 $\pm$ 0.006 \\
$k$-Dyck 5\% & 0.098 $\pm$ 0.004 & 0.135 $\pm$ 0.007 & 0.153 $\pm$ 0.007 \\
DeclRef 1\% & \textbf{0.097 $\pm$ 0.002} & 0.130 $\pm$ 0.014 & 0.152 $\pm$ 0.005 \\
DeclRef 5\% & \textbf{0.108 $\pm$ 0.011} & 0.125 $\pm$ 0.008 & 0.159 $\pm$ 0.005 \\
\midrule
Code 1\% (unified optimizer) & 0.098 $\pm$ 0.005 & 0.135 $\pm$ 0.002 & 0.159 $\pm$ 0.005 \\
Code 5\% (unified optimizer) & \textbf{0.103 $\pm$ 0.005} & 0.130 $\pm$ 0.008 & 0.158 $\pm$ 0.007 \\
Code 1\% (separate PPT optimizer) & \textbf{0.104 $\pm$ 0.004} & 0.132 $\pm$ 0.010 & 0.159 $\pm$ 0.006 \\
Code 5\% (separate PPT optimizer) & 0.096 $\pm$ 0.008 & 0.127 $\pm$ 0.012 & 0.156 $\pm$ 0.002 \\
\bottomrule
\end{tabular}

%% file: tables/constrained_bench.tex
\begin{tabular}{l cc}
\toprule
Method & BLiMP & DCLM CORE \\
\midrule
Baseline (0\% PPT) & 0.721 $\pm$ 0.004 & 0.068 $\pm$ 0.012 \\
NCA 1\% & 0.718 $\pm$ 0.003 & 0.077 $\pm$ 0.007 \\
NCA 5\% & 0.722 $\pm$ 0.001 & 0.074 $\pm$ 0.006 \\
NCA 10\% & 0.730 $\pm$ 0.006 & 0.069 $\pm$ 0.002 \\
$k$-Dyck 1\% & 0.714 $\pm$ 0.007 & 0.068 $\pm$ 0.009 \\
$k$-Dyck 5\% & 0.724 $\pm$ 0.001 & 0.059 $\pm$ 0.010 \\
$k$-Dyck 10\% & 0.728 $\pm$ 0.006 & 0.060 $\pm$ 0.008 \\
DeclRef 1\% & 0.727 $\pm$ 0.003 & 0.068 $\pm$ 0.007 \\
DeclRef 5\% & 0.720 $\pm$ 0.007 & 0.070 $\pm$ 0.009 \\
DeclRef 10\% & \textbf{0.731 $\pm$ 0.005} & 0.070 $\pm$ 0.004 \\
\bottomrule
\end{tabular}

%% file: tables/overchinchilla_bench.tex
\begin{tabular}{l cc}
\toprule
Method & BLiMP & DCLM CORE \\
\midrule
Baseline (0\% PPT) & 0.809 $\pm$ 0.006 & 0.263 $\pm$ 0.008 \\
NCA 1\% & 0.812 $\pm$ 0.005 & 0.264 $\pm$ 0.009 \\
$k$-Dyck 1\% & 0.809 $\pm$ 0.002 & 0.262 $\pm$ 0.009 \\
\bottomrule
\end{tabular}